%% file: main.tex
\documentclass[10pt,twocolumn,letterpaper]{article}

\usepackage{cvpr}
\usepackage{times}
\usepackage{epsfig}
\usepackage{graphicx}
\usepackage{amsmath}
\usepackage{amssymb}
\usepackage{amsfonts}
\usepackage{url}
\usepackage{appendix}

\usepackage{subcaption}
\usepackage[pagebackref=true,breaklinks=true,letterpaper=true,colorlinks,bookmarks=false]{hyperref}

\cvprfinalcopy % *** Uncomment this line for the final submission

\def\cvprPaperID{7072} % *** Enter the CVPR Paper ID here

\DeclareMathOperator{\lat}{z}

\ifcvprfinal\fi
\begin{document}

%%%%%%%%% TITLE
\title{Learning Predictive Models From Observation and Interaction}
%%CF.10.30: Here are some title ideas in addition to our original idea:
% Learning to Predict from Other Agents
% Action-Conditioned Prediction from Multiple Actors
% Action-Conditioned Prediction from Oneself and Others
% Learning to Predict by Observation
% Prediction By Observation and Interaction
% Learning World Models by Prediction and Interaction
%%SL.11.11: How about: Predicting From Observation and Interaction

%%CF: I editted the author list to be a different format. Feel free to revive the older version if you don't like it
\author{Karl Schmeckpeper$^1$, Annie Xie$^2$, Oleh Rybkin$^1$, Stephen Tian$^3$, \\Kostas Daniilidis$^1$, Sergey Levine$^3$, Chelsea Finn$^2$\\
$^1$University of Pennsylvania, $^2$Stanford University, $^3$UC Berkeley
%\\ 
%{\tt\small karls@seas.upenn.edu}
}

\iffalse
\author{Karl Schmeckpeper\\
University of Pennsylvania\\
Philadelphia, PA\\
{\tt\small karls@seas.upenn.edu}
\and
Annie Xie\\
Stanford University\\
Stanford, CA\\
{\tt\small anniexie@stanford.edu}
\and
Oleh Rybkin\\
University of Pennsylvania\\
Philadelphia, PA\\
{\tt\small oleh@seas.upenn.edu}
\and
Stephen Tian\\
UC Berkeley\\
Berkeley, CA\\
{\tt\small stephentian@berkeley.edu}
\and
Kostas Daniilidis\\
University of Pennsylvania\\
Philadelphia, PA\\
{\tt\small kostas@cis.upenn.edu}
\and
Sergey Levine\\
UC Berkeley\\
Berkeley, CA\\
{\tt\small svlevine@eecs.berkeley.edu}
\and
Chelsea Finn\\
Stanford University\\
Stanford, CA\\
{\tt\small cbfinn@cs.stanford.edu}
}
\fi

\maketitle
%\thispagestyle{empty}

%%%%%%%%% ABSTRACT
\begin{abstract}
   \input{sections/abstract.tex}
   
\end{abstract}

%%%%%%%%% BODY TEXT

\input{sections/introduction.tex}
\input{sections/related_work.tex}

\input{sections/methodology.tex}

\input{sections/experiments.tex}
\input{sections/conclusion.tex}

\input{sections/acknowledgements.tex}

{\small
\bibliographystyle{ieee_fullname}
\bibliography{definitions, bibtex}

}
\input{sections/appendix.tex}

\end{document}

%% file: sections/abstract.tex
Learning predictive models from interaction with the world allows an agent, such as a robot, to learn about how the world works, and then use this learned model to plan coordinated sequences of actions to bring about desired outcomes.
However, learning a model that captures the dynamics of complex skills represents a major challenge: if the agent needs a good model to perform these skills, it might never be able to collect the experience on its own that is required to learn these delicate and complex behaviors. Instead, we can imagine augmenting the training set with observational data of other agents, such as humans. Such data is likely more plentiful, but represents a different embodiment. For example, videos of humans might show a robot how to use a tool, but (i) are not annotated with suitable robot actions, and (ii) contain a systematic distributional shift due to the embodiment differences between humans and robots.
We address the first challenge by formulating the corresponding graphical model and treating the action as an observed variable for the interaction data and an unobserved variable for the observation data, and the second challenge by using a domain-dependent prior. In addition to interaction data, our method is able to leverage videos of passive observations in a driving dataset and a dataset of robotic manipulation videos. A robotic planning agent equipped with our method can learn to use tools in a tabletop robotic manipulation setting by observing humans without ever seeing a robotic video of tool use.

\footnotetext[1]{Correspondence to: Karl Schmeckpeper $<$karls@seas.upenn.edu$>$.}

%% file: sections/introduction.tex
\section{Introduction}

Humans have the ability to learn skills not just from their own interaction with the world but also by observing others. Consider an infant learning to use tools. In order to use a tool successfully, it needs to learn how the tool can interact with other objects, as well as how to move the tool to trigger this interaction. Such intuitive notion of physics can be learned by observing how adults use tools. More generally, observation is a powerful source of information about the world and how actions lead to outcomes. 
However, in the presence of physical differences (such as between an adult body and infant body), leveraging observation is challenging, as there is no direct correspondence between the demonstrator's and observer's actions. Evidence from neuroscience suggests that humans can effectively infer such correspondences and use it to learn from observation~\cite{rizzolatti_premotor_1996, rizzolatti2004mirror}. In this paper, we consider this problem: can we enable agents to learn to solve tasks using both their own interaction and the passive observation of other agents?

\begin{figure}[t]
\begin{center}
\includegraphics[width=0.95\linewidth]{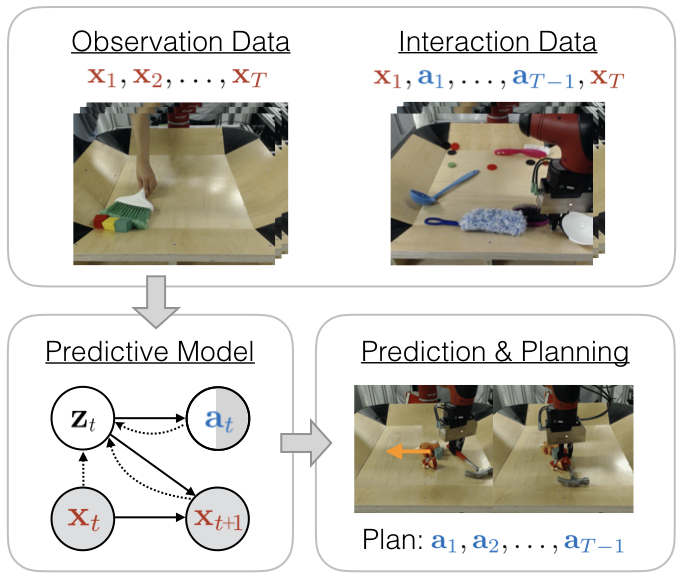}
\end{center}
\vspace{-0.3cm}
\caption{Our system learns from action-observation sequences collected through interaction, such as robotic manipulation or autonomous vehicle data, as well as observations of another demonstrator agent, such as data from a human or a dashboard camera. By combining interaction and observation data, our model is able to learn to generate predictions for complex tasks and new environments without costly expert demonstrations.}
\end{figure}

In model-based reinforcement learning, solving tasks is commonly addressed via learning action-conditioned predictive models. 
However, prior works have learned such predictive models from interaction data alone~\cite{hafner2018learning, Ha2018WorldModels, janner2019trust, ebert2018visual, Xie2019}. When using both interaction and observation data, the setup differs in two important ways. First, the actions of the observed agent are not known, and therefore directly learning an action-conditioned predictive model is not possible. Second, the observation data might suffer from a domain shift if the observed agent has a different embodiment, operates at a different skill level, or exists in a different environment.
Yet, if we can overcome these differences and effectively leverage observational data, we may be able to unlock a substantial source of broad data containing diverse behaviors and interactions with the world.

The main contribution of this work is an approach for learning predictive models that can leverage both videos of an agent annotated with actions and observational data for which actions are not available. We formulate a latent variable model for prediction, in which the actions are observed variables in the first case and unobserved variables in the second case.
We further address the domain shift between the observation and interaction data by learning a domain-specific prior over the latent variables. We instantiate the model with deep neural networks and train it with amortized variational inference.
In two problem settings -- driving and object manipulation -- we find that our method is able to effectively leverage observational data from dashboard cameras and humans, respectively, to improve the performance of action-conditioned prediction.
Further, we find that the resulting model enables a robot to solve tool-use tasks, and achieves significantly greater success than a model that does not use observational data of a human using tools.
Finally, we release our dataset of human demonstrations of tool use tasks to allow others to study this problem.

%% file: sections/related_work.tex
\section{Related Work}

\paragraph{Predictive models}

Video prediction can be used to learn useful representations and models in a fully unsupervised manner.
These representations can be used for tasks such as action recognition \cite{srivastava2015unsupervised},
action prediction \cite{vondrick2016anticipating},
classification \cite{denton2017unsupervised},
and planning \cite{finn2016unsupervised, finn2017deep, ebert2018visual, Janner2018, byravan2017se3, Ha2018WorldModels, hafner2018learning, kaiser2019modelbased, Fragkiadaki2016}.
Many different approaches have been applied to video prediction, including patch-centric methods~\cite{ranzato2014video}, compositional models of content and motion~\cite{villegas2017decomposing, denton2017unsupervised, tulyakov2017mocogan}, pixel autoregressive models~\cite{Kalchbrenner2016}, hierarchical models~\cite{Castrejon2019, mathieu2016, Luc2017}, transformation-based methods~\cite{Lotter2016, Patraucean2015, finn2016unsupervised, Vondrick2017, Liang2017, liu2017video, lee2018savp, babaeizadeh2018stochastic, Chen2017, VanAmersfoort2017}, and other techniques~\cite{DeBrabandere2016,wichers2018hierarchical,Byeon_2018_ECCV, ChaochaoLu2017}. We choose to leverage transformation-based models, as they have demonstrated good results on robotic control domains \cite{finn2016unsupervised, ebert2018visual}. Recent work has also developed stochastic video prediction models for better handling of uncertainty~\cite{denton2018stochastic, lee2018savp, babaeizadeh2018stochastic, Xue2016, Chen2017, Walker2016}. We also use a stochastic latent variable, and unlike these prior works, use it to model actions.

Learning action-conditioned visual dynamics models was proposed in \cite{Oh:2015:AVP:2969442.2969560, finn2016unsupervised, ChiappaRWM17}. Using MPC techniques and flow based prediction models, it has been applied to robotic manipulation \cite{finn2016unsupervised, finn2017deep, ebert2018visual, Janner2018, byravan2017se3, Zhang2018}. Other works address video games or physical simulation domains \cite{Ha2018WorldModels, hafner2018learning, kaiser2019modelbased, Fragkiadaki2016, Watter:2015:ECL:2969442.2969546}. 

\cite{Wang2019,Dasari2019, Yen-Chen2019} show these models can generalize to unseen tasks and objects while allowing for challenging manipulation of deformable objects, such as rope or clothing. Unfortunately, large amounts of robotic interaction data containing complex behavior are required to train these models. These models are unable to learn from cheap and abundantly available natural videos of humans as they are trained in \textit{action-conditioned} way, requiring corresponding control data for every video. 
In contrast, our method can learn from videos without actions, allowing it to leverage videos of agents for which the actions are unknown.

\paragraph{Learning to control without actions}

Recent work in imitation learning allows the agent to learn without access to the ground-truth expert actions.  
One set of approaches learn to translate the states of the expert into actions the agent can execute \cite{Torabi2018, Yu2018}.
\cite{DBLP:journals/corr/abs-1905-12612} uses action-free data to learn a set of sub-goals for hierarchical RL.
Another common approach is to learn a policy in the agent's domain that matches the expert trajectories under some similarity metric.  
\cite{Torabi2018a, Torabi2019, stadie2017third, Sun2019a} use adversarial training or other metrics to minimize the difference between the states generated by the demonstrated policy and the states generated by the learned policy.
\cite{Liu2018} transform images from the expert demonstrations into the robot's domain to make calculating the similarity between states generated by different policies in different environments more tractable.
\cite{Edwards2019} learn a latent policy on action-free data and use action-conditioned data to map the latent policy to real actions.
\cite{Sermanet2017TCN, dwibedi2018learning, NIPS2018_7557} learn state representations that can be used to transfer policies from humans to robots.
\cite{Sun2019} use partially action-conditioned data to train a generative adversarial network to synthesize the missing action sequences.
Unlike these works, which aim to specify a specific task to be solved through expert demonstrations, we aim to learn predictive models that can be used for multiple tasks, as we learn general properties of the real world through model-building.

Our work is more similar to a recently proposed method for learning action-conditioned predictive models without actions through learning action representations \cite{rybkin2019learning}. This work shows that very few active data points are required to learn sensorimotor mappings between true actions and action representations learned from observation in simple simulated settings. 
However, this approach addresses learning from observation where there is no physical differences between the demonstrator and the observer, and thus cannot be directly used for learning from humans.
Our approach explicitly considers domain shift, allowing it to leverage videos of humans to significantly outperform this approach.

\paragraph{Domain adaptation}
In order to handle both observational and interaction data, our method must handle the missing actions and bridge the gap between the two domains (e.g., human arms vs. robot arms). Related domain adaptation methods have sought to map samples in one domain into equivalent samples in another domain \cite{CycleGAN2017, Bousmalis2017, Taigman2017, Hoffman2018}, or learn feature embeddings with domain invariance losses \cite{Tzeng2014,Zhuang2015,ganin2015,Ganin2016,Tzeng2017}. In our setting, regularizing for invariance across domains is insufficient. For example, if the observational data of humans involves complex manipulation (e.g., tool use), while the interaction data involves only simple manipulation, we do not want the model to be invariant to these differences. We therefore take a different approach: instead of regularizing for invariance across domains, we explicitly model the distributions over (latent) action variables in each of the domains.

Related to our method, DIVA \cite{Ilse2019} aims to avoid losing this information by proposing a generative model with a partitioned latent space. The latent space is composed of both components  that are domain invariant and components that are conditioned on the domain. This allows the model to use domain-specific information while still remaining robust to domain shifts. We find that using an approach similar to DIVA in our model for learning from observation and interaction makes it more robust to the domain shift between interaction and observation data. However, in contrast to DIVA, our method explicitly handles sequence data with missing actions in one of the domains.

%% file: sections/methodology.tex
\newcommand{\x}{\textbf{x}}
\newcommand{\ac}{\textbf{a}}
\newcommand{\z}{\textbf{z}}

\section{Learning Predictive Models from Observation and Interaction}
In our problem setting, we assume access to interaction data of the form $\left[ \x_1, \ac_1, \dots, \ac_{T-1}, \x_T \right]$ and observation data of the form $\left[ \x_1, \dots, \x_T \right]$, where $\x_i$ denotes the $i^{\text{th}}$ frame of a video and $\ac_i$ denotes the action taken at the $i^{\text{th}}$ time step. Domain shift may exist between the two datasets:
for example, when learning object manipulation from videos of humans and robotic interaction, as considered in our experiments, there is a shift in the embodiment of the agent.
Within this problem setting, our goal is to learn an action-conditioned video prediction model, $p(\x_{c+1:T} | \x_{1:c}, \ac_{1:T})$, that predicts future frames conditioned on a set of $c$ context frames and sequence of actions.

To approach this problem, we formulate a probabilistic graphical model underlying the problem setting where actions are only observed in a subset of the data. In particular, in Subsection~\ref{subsection:graphical_model}, we introduce a latent variable that explains the transition from the current frame to the next and, in the case of interaction data, encodes the action taken by the agent. We further detail how the latent variable model is learned from both observation and interaction data by amortized variational inference. In Subsection~\ref{subsection:domain_shift}, we discuss how we handle domain shift by allowing the latent variables from different datasets to have different prior distributions. Finally, in Subsection~\ref{subsection:implementation}, we discuss specific implementation details of our model.

\subsection{Graphical Model}
\label{subsection:graphical_model}

\begin{figure}
\begin{center}
\includegraphics[width=0.85\linewidth]{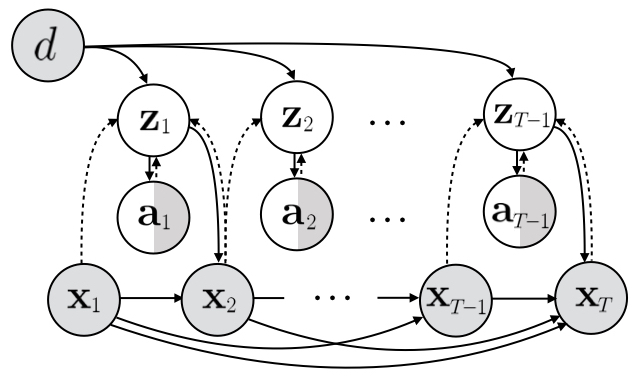}
\end{center}
\caption{\small We learn a predictive model of visual dynamics (in solid lines) that predicts the next frame $x_{t+1}$ conditioned on the current frame $x_t$ and action representation $z_t$. We optimize the likelihood of the interaction data, for which the actions are available, and observation data, for which the actions are missing. Our model is able to leverage joint training on the two kinds of data by learning a latent representation $z$ that corresponds to the true action. 
}
\label{fig:graphical_model}
\end{figure}

To leverage both passive observations and active interactions, we formulate the probabilistic graphical model depicted in Figure \ref{fig:graphical_model}. To model the action of the agent $\ac_t$, we introduce a latent variable $\z_t$, distributed according to a domain-dependent distribution. The latent $\z_{t}$ generates the action $\ac_{t}$.  
We further introduce a forward dynamic model that, at each time step $t$, generates the frame $\x_t$ given the previous frames $\x_{1:t-1}$ and latent variables $\z_{1:t-1}$.

The generative model can be summarized as:
\begin{align}
\z_t &\sim p(\z_t | d) \\
\ac_t &\sim p(\ac_t | \z_t) \\
\x_{t+1} &\sim p(\x_{t+1} | \x_{1:t}, \z_{1:t}).
\end{align}
The domain-dependent distribution over $\z_t$ is Gaussian with learned mean and variance, described in more detail in Subsection~\ref{subsection:domain_shift}, while the \textit{action decoder} $p(\ac_t | \z_t)$ and \textit{transition model} $p(\x_{t+1} | \x_{1:t}, \z_{1:t})$ are neural networks with Gaussian distribution outputs, described in Subsection~\ref{subsection:implementation}.

The transition model takes $\z_t$ as input and thus necessitates the posterior distributions $p(\z_t | \ac_t)$ and $p(\z_t | \x_t, \x_{t+1})$. We require $p(\z_t | \ac_t)$ to generate latent variables for action-conditioned video prediction, i.e. sampling from $p(\x_{t+1} | \x_{1:t}, \ac_{1:t}) = \mathbb{E}_{p(\z_{1:t} | \ac_{1:t})} \left[p(\x_{t+1} | \x_{1:t}, \z_{1:t})\right]$.
We also require $p(\z_t | \x_t, \x_{t+1})$
since the actions are not available in some trajectories to obtain the first distribution.

The computation of these two posterior distributions is intractable, since the model is highly complex and non-linear, so we introduce the variational distributions $q_{\text{act}}(\z_t | \ac_t)$ and $q_{\text{inv}}(\z_t | \x_t, \x_{t+1})$ to approximate $p(\z_t | \ac_t)$ and $p(\z_t | \x_t, \x_{t+1})$. The distributions are modeled as Gaussian and the variational parameters are learned by optimizing the evidence lower bound (ELBO),
which is constructed by considering two separate cases. In the first, the actions of a trajectory are observed, and we optimize an ELBO on the joint probability of the frames and the actions:
\begin{equation}
    \begin{aligned}
    \log p(\x_{1:T}, \ac_{1:T}) &\ge \mathbb{E}_{q_{\text{act}}(\z_{1:t} | \ac_{1:t})} \left[ \sum_t \log p(\x_{t+1} | \x_{1:t}, \z_{1:t}) \right] \\
    &~~~~+ \mathbb{E}_{q_{\text{act}}(\z_{t} | \ac_{t})} \left[ \sum_t \log p(\ac_t | \z_t) \right]\\
    &~~~~- \sum_t D_{KL} (q_{\text{act}}(\z_{t} | \ac_{t}) || p(\z_{t})) \\
    &= -\mathcal{L}_{i}(\x_{1:T}, \ac_{1:T}).
    \end{aligned}
    \label{eqn:variatonal_lower_bound_observerd}
\end{equation}

In the second case, the actions are not observed, and we optimize an ELBO on only the probability of the frames:
\begin{equation}
    \begin{aligned}
    \log p(\x_{1:T}) &\ge  \mathbb{E}_{q_{\text{inv}}(\z_t | \x_t, \x_{t+1})} \left[ \sum_t \log p(\x_{t+1} | \x_t, \z_t) \right] \\
    &~~~~- \sum_t D_{KL} (q_{\text{inv}}(\z_t | \x_t, \x_{t+1}) || p(\z_t)) \\
    &= -\mathcal{L}_{o}(\x_{1:T}).
    \end{aligned}
    \label{eqn:variational_lower_bound_unobserved}
\end{equation}

The ELBO for the entire dataset is the combination of the lower bounds for the interaction data with actions, $D^i$, and the observation data without actions, $D^o$:
\begin{equation}
    \mathcal{J} = \sum_{(\x_{1:T}, \ac_{1:T}) \sim D^i} \mathcal{L}_i(\x_{1:T}, \ac_{1:T}) + \sum_{\x_{1:T} \sim D^o} \mathcal{L}_o (\x_{1:T}).
    \label{eqn:combined_variational_lower_bound}
\end{equation}

We also add an auxiliary loss to align the distributions of $\z$ generated from the encoders $q_{\text{act}}(\z_t | \ac_t)$ and $q_{\text{inv}}(\z_t | \x_t, \x_{t+1})$, since the encoding $\z$ should be independent of the distribution it was sampled from. We encourage the two distributions to be similar through the Jensen-Shannon divergence:
\begin{equation}
    \mathcal{L}_{JS} = \sum_{(\x_{1:T}, \ac_{1:T}) \sim D^i} D_{JS}(q_{\text{act}}(\z_t | \ac_t) \| q_{\text{inv}}(\z_t | \x_t, \x_{t+1})).
    \label{eqn:jensen_shannon}
\end{equation}

Our final objective combines the evidence lower bound for the entire dataset and the Jensen-Shannon divergence, computed for the interaction data:
\begin{equation}
    \mathcal{F} = \mathcal{J} + \alpha \mathcal{L}_{JS}.
    \label{eqn:combined_objective}
\end{equation}

We refer to our approach as prediction from observation and interaction (POI).

\subsection{Domain Shift}
\label{subsection:domain_shift}
When learning from both observation and interaction, domain shift may exist between the two datasets. For instance, in the case of a robot learning by observing people, the two agents differ both in their physical appearance, as well as their action spaces. To address these domain shifts,
we take inspiration from the domain-invariant approach described in \cite{Ilse2019}.  We divide our latent variable $\z$ into $\z^{\text{shared}}$, which captures the parts of the latent action that are shared between domains, and $\z^{\text{domain}}$, which captures the parts of the latent action that are unique to each domain.  

We allow the network to learn the difference between the $\z^{\text{domain}}$ for each dataset by using different prior distributions. The prior $p(\z^{\text{shared}}_t)$ is the same for both domains, however, the prior for $\z^{\text{domain}}_t$ is different for the interaction dataset, $p_{i}(\z^{\text{domain}}_t)$, and the observational dataset, $p_{o}(\z^{\text{domain}}_t)$. 
$p(\z^{\text{shared}}_t)$ and $p_{a}(\z^{\text{domain}}_t)$ are both multivariate Gaussian distributions with a learned mean and variance for each dimension. The prior is the same for all timesteps $t$.

Unlike the actions for the robot data, which are sampled from the same distribution at each time step, the actions of the human are correlated across time.
For the human observation data, the prior $p_{o}(\z^{\text{domain}}_{1:T} | \x_1)$ models a joint distribution over timesteps, and
is parameterized as a long short-term memory (LSTM) network \cite{hochreiter1997long}. The input to the LSTM at the first timestep is an encoding of the initial observation, and the LSTM cell produces the parameters of the multivariate Gaussian distribution for each time step.

\begin{figure}
\begin{center}
\includegraphics[width=\linewidth]{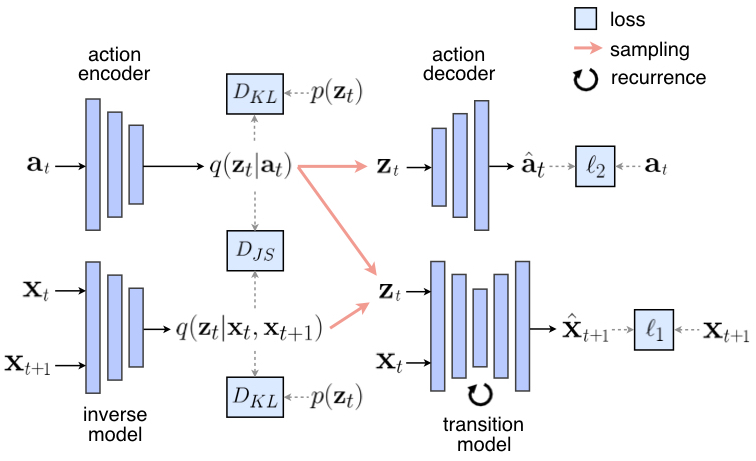}
\end{center}
\vspace{-0.4cm}
\caption{\small Network architecture. To optimize the ELBO, we predict the latent action ${\z}_t$ from $\x_t$ and $\x_{t+1}$ using the inverse model $q_\text{inv}$. When the true actions are available, we additionally predict the latent action from the true action $\ac_t$ using the action encoder $q_\text{act}$, and encourage the predictions from $q_\text{act}$ and $q_\text{inv}$ to be similar with a Jensen-Shannon divergence loss. The next frame is predicted from $\z_t$ and $\x_t$. 
}
\label{fig:network_diagram}
\end{figure}

\subsection{Deep Neural Network Implementation}
\label{subsection:implementation}

A high-level diagram of our network architecture is shown in Figure \ref{fig:network_diagram}, and a more detailed version is presented in Appendix~\ref{appendix:full_architecture}. 
Our action encoder $q_{\text{act}}(\z_t | \ac_t)$ is a multi-layer perceptron with 3 layers of 64 units to encode the given action $\ac_t$ to the means and variances for each dimension of the encoding.  

Our inverse model $q_{\text{inv}}(\z_t | \x_t, \x_{t+1})$
is a convolutional network that predicts the distribution over the action encoding.  The network is made up of three convolutional layers with \{32, 64, 128\} features with a kernel size of 4 and a stride of 2.  Each convolutional layer is followed by instance normalization and a leaky-ReLU.  The output of the final convolutional layer is fed in a fully connected layer, which predicts the means and variances of the action encoding.

We encourage the action encodings generated by the action encoder $q_{\text{act}}$ and the inverse model $q_{\text{inv}}$ to be similar using the Jensen-Shannon divergence in Equation~\ref{eqn:jensen_shannon}.
Since the Jensen-Shannon divergence does not have a closed form solution, we approximate it by using a mean of the Gaussians instead of a mixture.  
Our model uses a modified version of the SAVP architecture \cite{lee2018savp} as the transition model which predicts $\x_{t+1}$ from $\x_t$ and an action encoding $\z$, either sampled from $q_{\text{act}}(\z_t | \ac_t)$ or from $q_{\text{inv}}(\z_t | \x_t, \x_{t+1})$. In the case where the actions are observed, we generate two predictions, one from each of $q_{\text{inv}}$ and $q_{\text{act}}$, and in the case where actions are not observed, we only generate a prediction from the inverse model, $q_{\text{inv}}$. This architecture has been shown to be a useful transition model for robotic planning in \cite{ebert2018visual, Dasari2019}.

Our action decoder predicts the mean of the distribution $p(\ac_t | \z_t)$ using a multi-layer perceptron with 3 layers of 64 units each, while using a fixed unit variance.

%% file: sections/experiments.tex
\section{Experiments}
\noindent We aim to answer the following in our experiments: 
\vspace{-0.2cm}
\begin{enumerate}
    \item Do passive observations, when utilized effectively, improve an action-conditioned visual predictive model despite large domain shifts?
    \vspace{-0.2cm}
    \item How does our approach compare to alternative methods for combining passive and interaction data?
    \vspace{-0.2cm}
    \item Do improvements in the model transfer to downstream tasks, such as robotic control?
\end{enumerate}
\vspace{-0.2cm}
To answer 1, we compare our method to a strong action-conditioned prediction baseline, SAVP \cite{lee2018savp}, which is trained only on interaction data as it is not able to leverage the observation data. To answer 2, we further compare to a prior method for inferring actions from action-free data, CLASP \cite{rybkin2019learning}, and a method that imputes missing actions based on a shared inverse model, described below. 
We study questions 1 and 2 in both the driving domain in Subsection~\ref{subsection:nuscenes} and the robotic manipulation domain in Subsection~\ref{subsection:robot_prediction} and evaluate the methods on action-conditioned prediction. We evaluate question 3 by directly controlling the robotic manipulator using our learned model. Videos of our results are available on the supplementary website\footnote{Our supplementary website is at \url{https://sites.google.com/view/lpmfoai}}.

As an additional point of comparison, we propose a shared inverse model that draws similarities to label propagation in semi-supervised learning~\cite{label_prop}.
In this comparison, an inverse model and transition model are jointly learned on all of the data. Specifically, the inverse model predicts the action taken between a pair of images, supervised only by the actions from the interaction data, and the transition model predicts the next frame conditioned on the current frame and an input action.
When available, the transition model uses the true action, otherwise, it uses the action predicted by the inverse model.

\subsection{Visual Prediction for Driving}
\label{subsection:nuscenes}

\begin{table*}
\small
\begin{center}
\begin{tabular}{|l|c|c|c|}
\hline
Method & PSNR ($\uparrow$) & SSIM ($\uparrow$) & LPIPS \cite{zhang2018perceptual} ($\downarrow$)\\
\hline\hline
SAVP \cite{lee2018savp} (Boston w/ actions)
& $19.74\pm0.41$ & $0.5121\pm0.0164$ & $0.1951\pm0.0075$ \\

Shared Inverse Model (Boston w/ actions, Singapore w/o actions)
& $20.65\pm0.52$ & $0.5455\pm0.0166$ & $0.2003\pm0.0080$ \\

CLASP \cite{rybkin2019learning} (Boston w/ actions, Singapore w/o actions) 
& $20.57\pm0.48$ & $0.5431\pm0.0161$ & $0.1964\pm0.0076$ \\

POI (ours) (Boston w/ actions, BDD100K w/o actions)
& $\mathbf{20.88\pm0.24}$ & $\mathbf{0.5508\pm0.0076}$ & $0.2106\pm0.0089$ \\

POI (ours) (Boston w/ actions, Singapore w/o actions) 
& $20.81\pm0.49$ & $0.5486\pm0.0164$ & $\mathbf{0.1933\pm0.0074}$ \\

\hline
Oracle - SAVP \cite{lee2018savp} (Boston w/ actions. Singapore w/ actions)
& $21.17\pm0.47$ & $0.5752\pm0.0156$ & $0.1738 \pm0.0076$\\

\hline
\end{tabular}
\end{center}
\vspace{-0.4cm}
\caption{\small Means and standard errors for action-conditioned prediction on the Singapore portion of the nuScenes dataset. By leveraging observational driving data from Singapore or from BDD dashboard cameras, our method is able to outperform prior models that cannot leverage such data (i.e. SAVP) and slightly outperform alternative approaches to using such data.}
\label{table:nuscenes_metrics}
\vspace{-0.4cm}
\end{table*}

\input{sections/diagrams/nuscenes_predictions.tex}

We first evaluate our model on video prediction for driving. Imagine that a self-driving car company has data from a fleet of cars with sensors that record both video and the driver's actions in one city, and a second fleet of cars that only record dashboard video, without actions, in a second city. If the goal is to train an action-conditioned model that can be utilized to predict the outcomes of steering actions, our method allows us to train such a model using data from both cities, even though only one of them has actions.

We use the nuScenes \cite{nuscenes2019} and BDD100K \cite{Yu2018a} datasets for our experiments.  The nuScenes dataset consists of 1000 driving sequences collected in either Boston or Singapore, while the BDD100K dataset contains only video from dashboard cameras.
In nuScenes, we discard all action and state information for the data collected in Singapore, simulating data that could have been collected by a car equipped with only a camera.  
We train our model with action-conditioned video from Boston and action-free video either from the nuScenes Singapore data or the BDD100K data, and evaluate on action-conditioned prediction on held-out data from Singapore (from nuScenes).
Since the action distribution for all datasets is likely very similar as they all contain human driving, we use the same learned means and variances for the Gaussian prior over $\lat$ for both portions of the dataset. 
We additionally train a our model with the action-conditioned video from Boston and action-free video taken from the BDD100K dataset \cite{Yu2018a}.

We compare our predictions to those generated by the SAVP \cite{lee2018savp} model trained with only the action-conditioned data from Boston, since SAVP cannot leverage action-free data for action-conditioned prediction.
We additionally compare our predictions to those generated by CLASP \cite{rybkin2019learning} and the shared inverse model, both trained with action-conditioned video from Boston, and action-free video from Singapore.
As an upper-bound, we train the SAVP \cite{lee2018savp} model with action-conditioned data from Boston and action-conditioned data from Singapore.

Comparisons between these methods are shown in Table~\ref{table:nuscenes_metrics}. 
Qualitative results are shown in Figure~\ref{fig:nuscenes_predictions}.
With either form of observational data, BDD2K or nuScenes Singapore, our method significantly outperforms the SAVP model trained with only action-conditioned data from Boston, demonstrating that our model can leverage observation data to improve the quality of its predictions. Further, our method slightly outperforms alternative approaches to learning from observation and interaction.

\subsection{Robotic Manipulation: Prediction}
\label{subsection:robot_prediction}
\input{sections/diagrams/dataset_examples.tex}

\input{sections/diagrams/robot_predictions.tex}
\begin{table*}
\begin{center}
\small
\begin{tabular}{|l|c|c|c|c|}
\hline
Method & PSNR ($\uparrow$) & SSIM ($\uparrow$) & LPIPS \cite{zhang2018perceptual} ($\downarrow$)\\
\hline\hline
CLASP \cite{rybkin2019learning} (random robot, expert human) 
& $22.14 \pm 0.11$ & $0.763 \pm 0.004$ & $0.0998\pm 0.0023$\\

SAVP \cite{lee2018savp} (random robot) 
& $23.31 \pm 0.10$ & $0.803 \pm 0.004$ & $0.0757 \pm 0.0022$\\

Shared IM (random robot, expert human)
& $23.59\pm0.10$ & $0.808\pm0.004$ & $0.0770\pm0.0022$\\

POI (ours) (random robot, expert human) 
& $\mathbf{23.79\pm0.12}$ & $\mathbf{0.813\pm0.005}$ & $\mathbf{0.0722\pm0.0024}$\\

\hline
Oracle \cite{lee2018savp} (random robot, expert kinesthetic) 
& $24.99 \pm 0.11$  & $0.858\pm0.003$ & $0.0486 \pm 0.0017$\\

\hline
\end{tabular}
\end{center}
\vspace{-0.5cm}
\caption{\small Means and standard errors for action-conditioned prediction on the manipulation dataset.  
By leveraging observational data of human tool use, our model was able to outperform prior models that cannot leverage such data (i.e. SAVP) and slightly outperform alternative approaches to using such data.}
\label{table:robotic_video_metrics}
\vspace{-0.3cm}
\end{table*}

\begin{figure}
    \centering
    \begin{subfigure}{0.3\linewidth}
    \centering
    \includegraphics[width=1.0\linewidth]{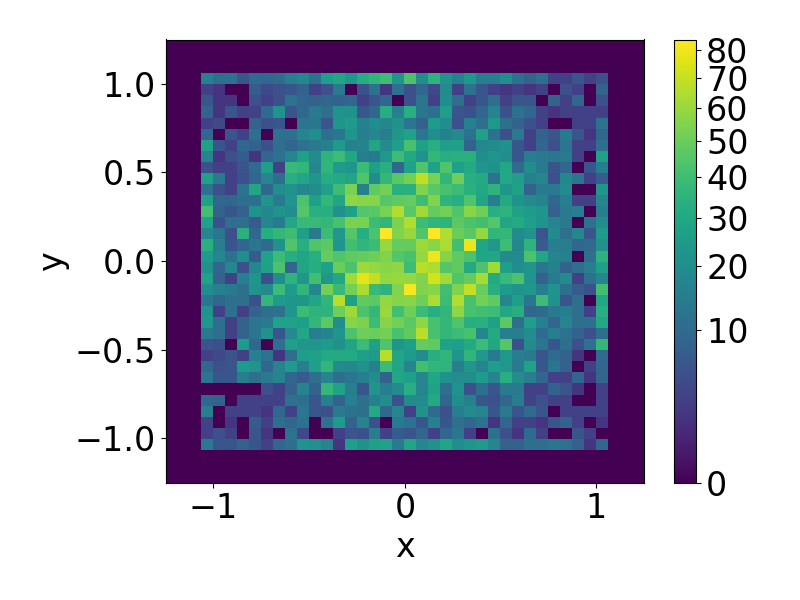}
    \caption{\centering \mbox{Random} \mbox{robot} \mbox{actions}}
    \end{subfigure}
    \begin{subfigure}{0.3\linewidth}
    \centering
    \includegraphics[width=\linewidth]{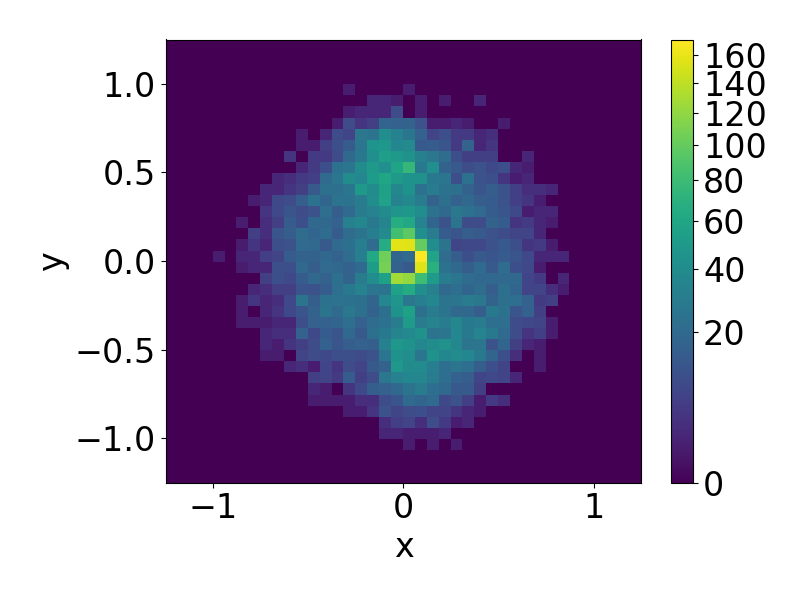}
    \caption{\centering Expert robot \mbox{actions}}
    \end{subfigure}
    \begin{subfigure}{0.3\linewidth}
    \centering
    \includegraphics[width=\linewidth]{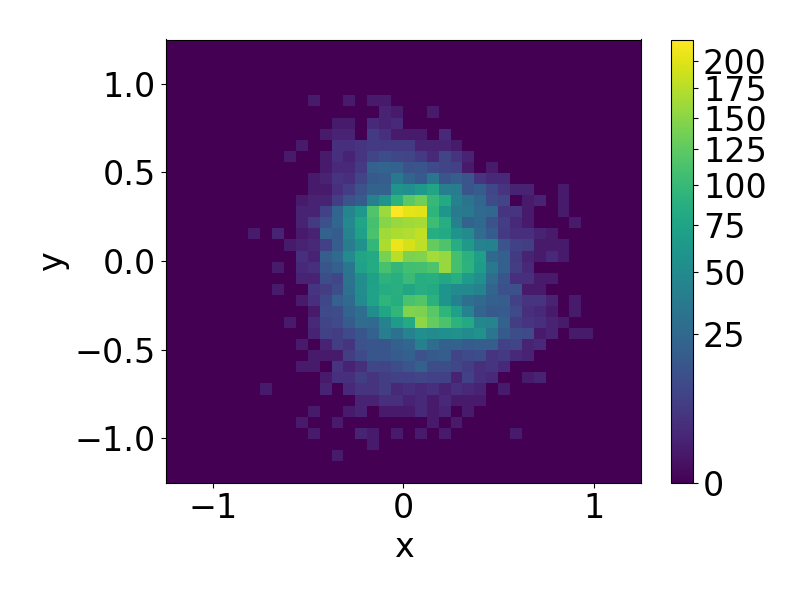}
    \caption{\centering Predicted \mbox{human} actions}
    \end{subfigure}
    \caption{\small Histograms of the x and y components of the actions.  Since the human data does not have any actions, the displayed actions were generated by our inverse model. The distribution of predicted human actions of tool-use resembles that of the expert robot actions, suggesting that our model has learned to successfully decode human actions.}
    \label{fig:action_hists_2d}
    \vspace{-0.3cm}
\end{figure}

\begin{figure*}
\begin{center}
\includegraphics[width=\textwidth]{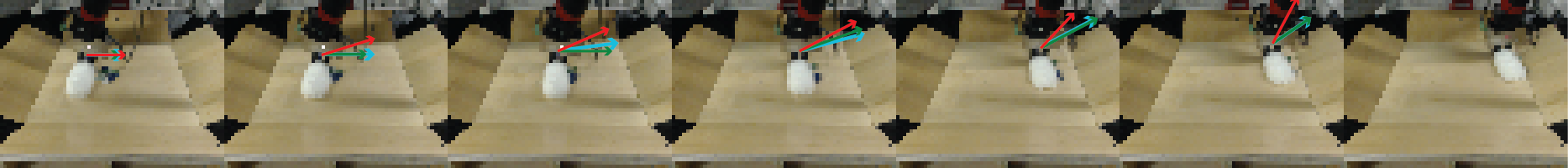}

\includegraphics[width=\textwidth]{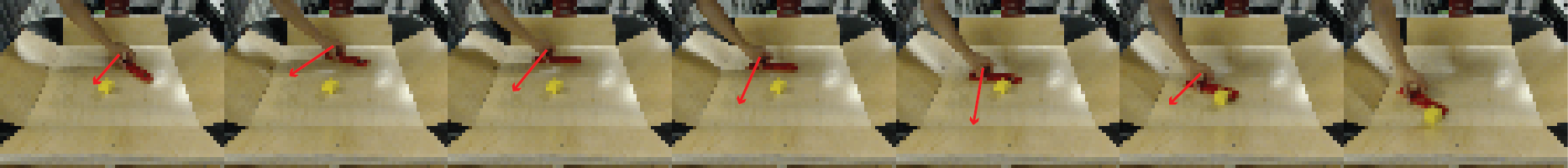}
\end{center}
\vspace{-0.4cm}
   \caption{\small Action predictions on human and robot data.  The sequences of images show the ground truth observations, while the arrows show the action in the (x, y) plane between each pair of frames.  The blue arrow is the ground truth action, the green arrow is the action generated from decoding the output of the action encoder, and the red is the action generated by decoding the output of the inverse model.  The human data only has actions generated by the inverse model. Our model is able to infer plausible actions for both domains, despite never seeing ground truth human actions.}
\label{fig:action_predictions}
\vspace{-0.3cm}
\end{figure*}

We evaluate our model on the robotic manipulation domain, which presents a large distributional shift challenge between robot and human videos. In particular, we study a tool-use task and evaluate whether human videos of tool-use can improve predictions of robotic tool-use interactions. 

For our interaction data, we acquired 20,000 random trajectories of a Sawyer robot from the open-source datasets from \cite{ebert2018visual} and \cite{Xie2019}, which consist of both video and corresponding actions. We then collected 1,000 videos of a human using different tools to push objects as the observation data. By including the human videos, we provide the model with examples of tool-use interactions, which are not available in the random robot data. Our test set is composed of 1,200 kinesthetic demonstrations from \cite{Xie2019}, in which a human guides the robot to use tools to complete pushing tasks similar to those in the human videos. Kinesthetic demonstrations are time-consuming to collect, encouraging us to build a system that can be trained without them, but they serve as a good proxy for evaluating robot tool-use behavior.
Example images from the datasets are shown in Figure~\ref{fig:dataset_examples}.
\footnote{All components of our dataset will be released upon publication.}
This dataset is especially challenging because of the large domain shift between the robot and human data.  The human arm has a different appearance from the robot and moves in a different action space.

We compare to the CLASP model \cite{rybkin2019learning} and a shared inverse model, both trained with the same data as our model.
We also evaluate the SAVP model \cite{lee2018savp}, trained the same robot data, but without the human data, since the SAVP model is unable to leverage action-free data for action-conditioned prediction.

For an oracle, we trained the SAVP model \cite{lee2018savp} on both the random robot trajectories and the kinesthetic demonstrations.

As shown in Table \ref{table:robotic_video_metrics}, our model is able to leverage information from the human videos to outperform the other models.  Both our model and the shared inverse model outperformed the SAVP model trained on only the random robot data, showing that it is possible to leverage passive observation data to improve action-conditioned prediction, even in the presence of the large domain shift between human and robot arms.
Our model also outperformed our shared inverse model, showing the importance of explicitly considering domain shift and stochasticity.

Qualitative results are shown in Figure \ref{fig:robotic_predictions}.  Our model is able to generate more accurate predictions than the baseline SAVP model that was trained with only the robotic interaction data.
In addition to predicting future states, our model is able to predict the action that occurred between two states.  Examples for both robot and human demonstrations are shown in Figure \ref{fig:action_predictions}. Our inverse model is able to generate reasonable actions for both the robot and the human data despite having never been trained on human data with actions.  
Histograms of the action distributions for different parts of our dataset are shown in Figure \ref{fig:action_hists_2d}.  Our model is able to extract actions for the human data from a reasonable distribution.  
Our model maps human and robot actions to a similar space, allowing it to exploit their similarities to improve prediction performance on robotic tasks.

\subsection{Robotic Manipulation: Planning and Control}
\label{subsection:robot_planning}
\input{sections/diagrams/robot_manipulation.tex}
To study the third and final research question, we further evaluate the efficacy of our visual dynamics model in a set of robotic control experiments. We combine our model with sampling-based visual model predictive control, which optimizes actions with respect to a user-provided task \cite{finn2017deep, ebert2018visual}. In each task setting, several objects, as well as a tool that the robot could potentially use to complete the task, are placed in the scene. Tasks are specified by designating a pixel corresponding to an object and the goal position for the object, following \cite{finn2017deep,ebert2018visual}. We specify moving multiple objects by selecting multiple pairs of pixels. 

To evaluate the importance of the human data, we focus on control tasks that involve moving multiple objects, which would be difficult to complete without using a tool. We quantitatively evaluate each model on 15 tasks with tools seen during training and 15 tasks with previously unseen tools. In Figure~\ref{fig:robotic_control_qual}, we show qualitative examples of the robot completing tool-use tasks.

The quantitative results, in Table~\ref{table:robotic_control_quant}, indicate that the planner can leverage our model to execute more successful plans relative to the baseline SAVP model, which was trained only using random robot trajectories. In our evaluation, a trial is successful if the average distance between the objects and their respective goal positions at the final time step is less than or equal to 10 centimeters.
Using our model, the robot achieves similar performance to the oracle model trained on kinesthetic demonstrations with action labels. This result suggests that our model has effectively learned about tool-object interactions by observing humans.

\begin{table}
\small
\begin{center}
\begin{tabular}{|l|c|}
\hline
Method & Success Rate \\
\hline\hline
SAVP \cite{lee2018savp} (random) & $23.3\pm7.7\%$ \\
POI (ours) (random, human) & $\mathbf{40.0\pm8.9\%}$ \\
\hline
Oracle (random, kinesthetic) & $36.7\pm8.8\%$ \\
\hline
\end{tabular}
\end{center}
\vspace{-0.4cm}
\caption{\small Success rates \& standard errors for robotic tasks.  "random" denotes random robot data, "human" denotes human interaction data, and "kinesthetic" is an oracle dataset of expert robot trajectories. Our model performs comparably to the oracle, and successfully leverages the observational videos to improve over SAVP.}
\label{table:robotic_control_quant}
\vspace{-0.4cm}
\end{table}

%% file: sections/diagrams/nuscenes_predictions.tex
\begin{figure}
\begin{center}

\rotatebox{90}{\scriptsize{Ground Truth}}
\includegraphics[width=0.26\linewidth]{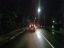}
\includegraphics[width=0.26\linewidth]{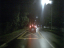}
\includegraphics[width=0.26\linewidth]{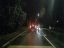}
\\
\rotatebox{90}{\scriptsize{Baseline}}
\includegraphics[width=0.26\linewidth]{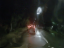}
\includegraphics[width=0.26\linewidth]{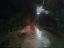}
\includegraphics[width=0.26\linewidth]{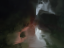}
\\
\rotatebox{90}{\scriptsize{POI (ours)}}
\includegraphics[width=0.26\linewidth]{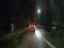}
\includegraphics[width=0.26\linewidth]{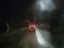}
\includegraphics[width=0.26\linewidth]{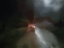}

\end{center}
\vspace{-0.4cm}

\newcommand{\labwidth}{0.184\linewidth}
\hspace{0.14\linewidth}
t = 0\hspace{\labwidth}
t = 2\hspace{\labwidth}
t = 4

   \caption{\small Example predictions on the Singapore portion of the Nuscenes dataset. This sequence was selected for large MSE difference between the models. More examples are available in the supplementary material.  We compare our model to the baseline of the SAVP model trained on the Boston data with actions. Our model is able to maintain the shape of the car in front.}
   \vspace{-0.3cm}
\label{fig:nuscenes_predictions}
\end{figure}

%% file: sections/diagrams/dataset_examples.tex
\begin{figure}
\newcommand{\predwidth}{0.26\linewidth}
\begin{center}

\includegraphics[width=\predwidth]{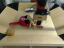}
\includegraphics[width=\predwidth]{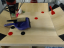}
\includegraphics[width=\predwidth]{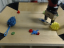}
\hspace{0.5cm}
\includegraphics[width=\predwidth]{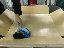}
\includegraphics[width=\predwidth]{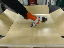}
\includegraphics[width=\predwidth]{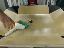}
\end{center}
\vspace{-0.4cm}
   \caption{\small Example images from the robot (top) and human (bottom) datasets.}
\label{fig:dataset_examples}
\vspace{-0.3cm}
\end{figure}

%% file: sections/diagrams/robot_predictions.tex
\begin{figure*}
\newcommand{\predwidth}{0.13\linewidth}
\begin{center}
\rotatebox{90}{\scriptsize{Ground Truth}}
\includegraphics[width=\predwidth]{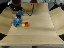}
\includegraphics[width=\predwidth]{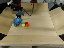}
\includegraphics[width=\predwidth]{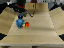}
\includegraphics[width=\predwidth]{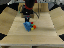}
\includegraphics[width=\predwidth]{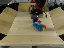}
\includegraphics[width=\predwidth]{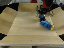}
\includegraphics[width=\predwidth]{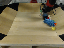}
\\
\rotatebox{90}{\scriptsize{Baseline}}
\includegraphics[width=\predwidth]{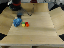}
\includegraphics[width=\predwidth]{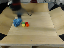}
\includegraphics[width=\predwidth]{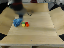}
\includegraphics[width=\predwidth]{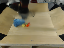}
\includegraphics[width=\predwidth]{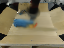}
\includegraphics[width=\predwidth]{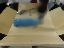}
\includegraphics[width=\predwidth]{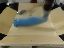}
\\
\rotatebox{90}{\scriptsize{POI (ours)}}
\includegraphics[width=\predwidth]{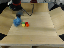}
\includegraphics[width=\predwidth]{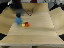}
\includegraphics[width=\predwidth]{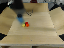}
\includegraphics[width=\predwidth]{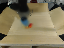}
\includegraphics[width=\predwidth]{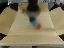}
\includegraphics[width=\predwidth]{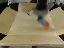}
\includegraphics[width=\predwidth]{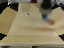}
\\
\end{center}
\vspace{-0.4cm}
\newcommand{\labwidth}{0.092\linewidth}
\hspace{0.07\linewidth}
t = 0\hspace{\labwidth}
t = 2\hspace{\labwidth}
t = 4\hspace{\labwidth}
t = 6\hspace{\labwidth}
t = 8\hspace{\labwidth}
t = 10\hspace{\labwidth}
t = 12

   \caption{\small Example predictions on the robotic dataset. We compare our model to the baseline of the SAVP model trained with random robot data. This sequence was selected to maximize the MSE difference between the models. More examples are available in the supplementary material. Our model more accurately predicts both the tool and the object it pushes.}
\label{fig:robotic_predictions}
\end{figure*}

%% file: sections/diagrams/robot_manipulation.tex
\newcommand{\imwidth}{0.13\linewidth}

\begin{figure*}
\begin{center}
\includegraphics[width=\imwidth]{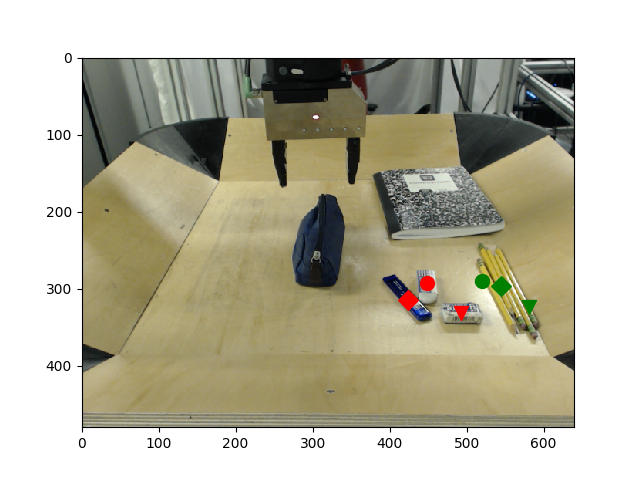}
\includegraphics[width=\imwidth]{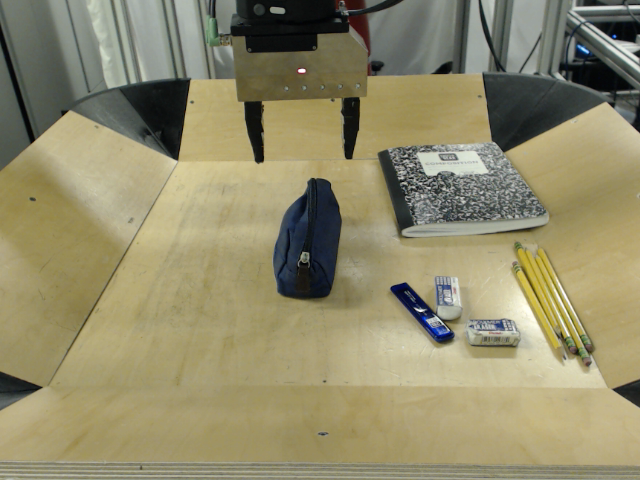}
\includegraphics[width=\imwidth]{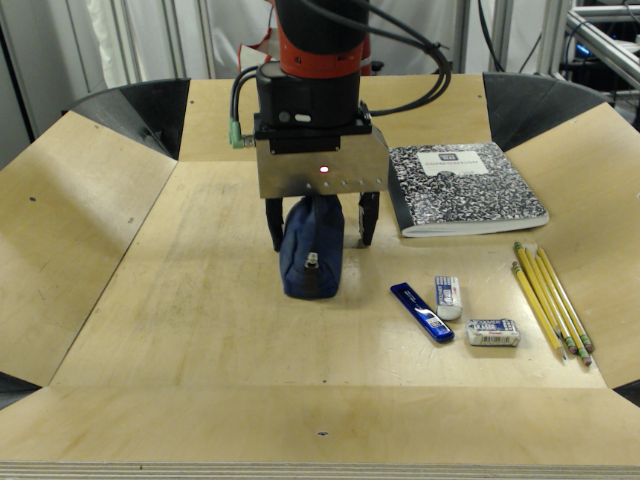}
\includegraphics[width=\imwidth]{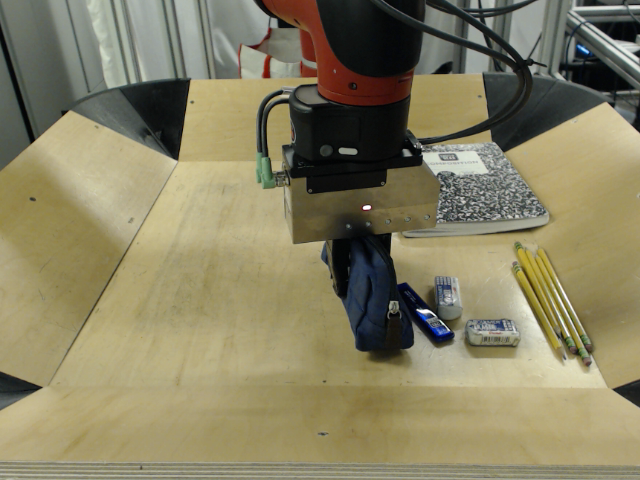}
\includegraphics[width=\imwidth]{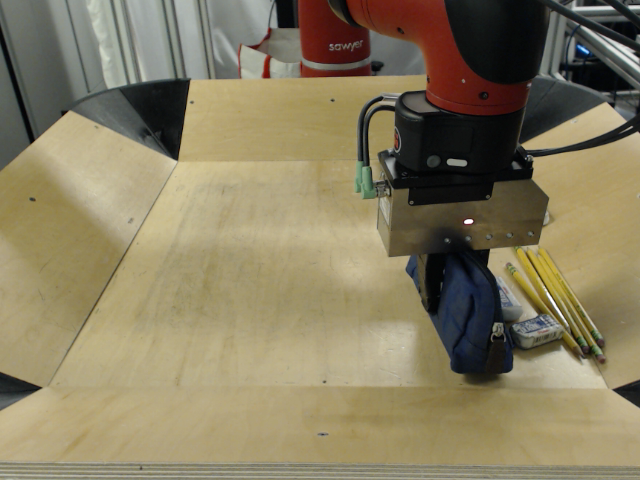}
\includegraphics[width=\imwidth]{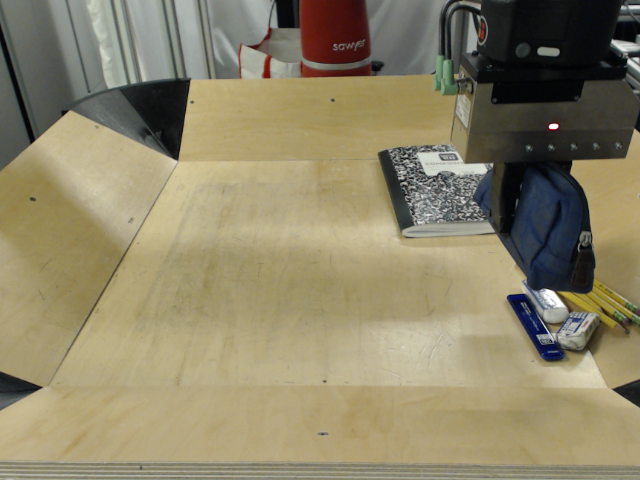}
\includegraphics[width=\imwidth]{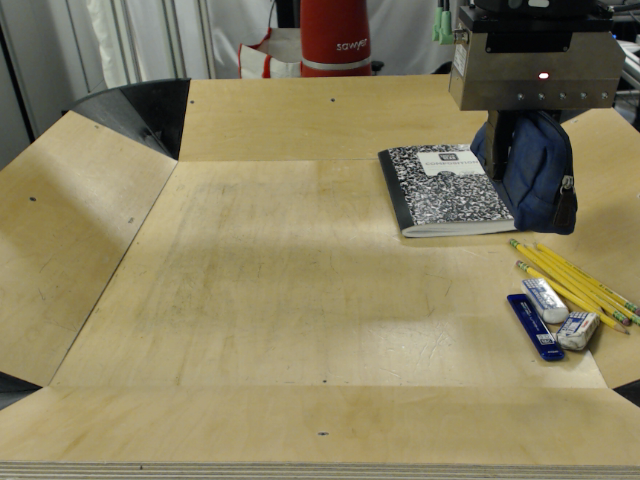}

\vspace{0.25cm}
\includegraphics[width=\imwidth]{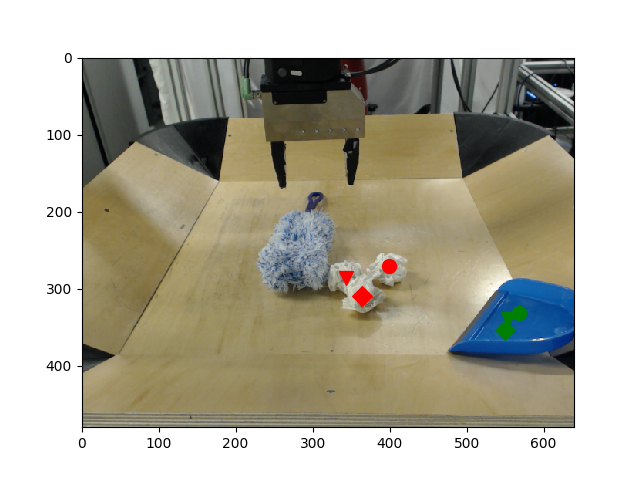}
\includegraphics[width=\imwidth]{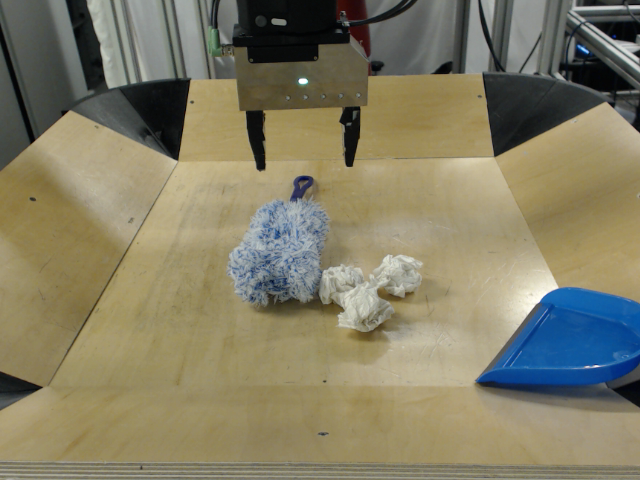}
\includegraphics[width=\imwidth]{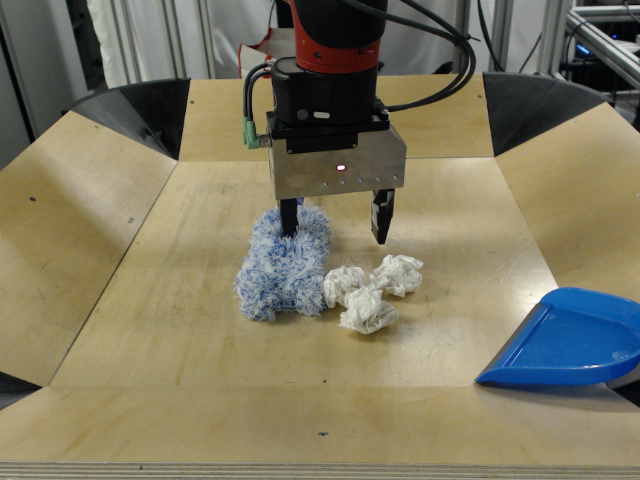}
\includegraphics[width=\imwidth]{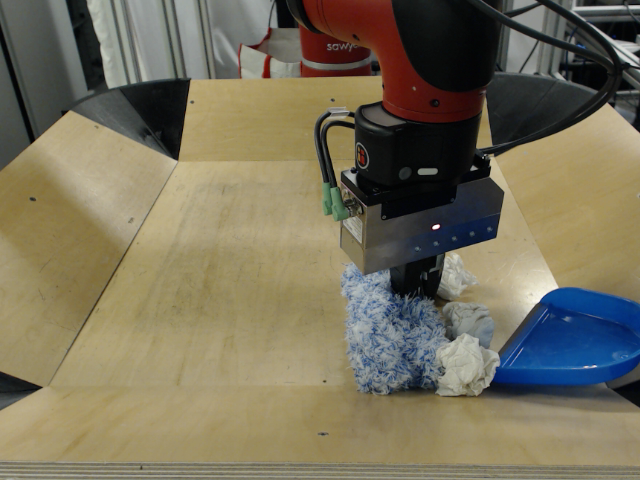}
\includegraphics[width=\imwidth]{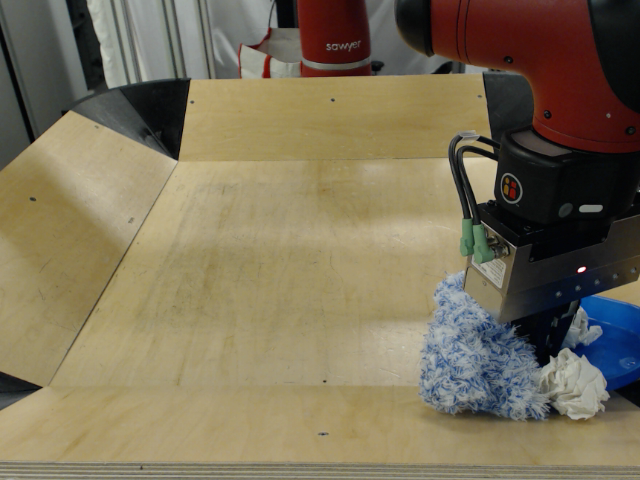}
\includegraphics[width=\imwidth]{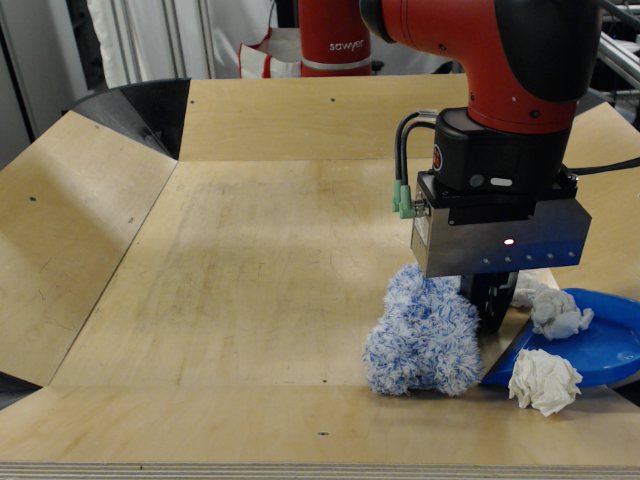}
\includegraphics[width=\imwidth]{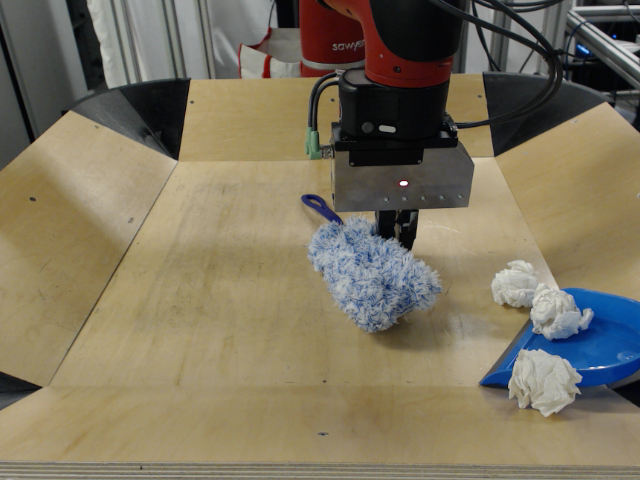}

\end{center}
\vspace{-0.4cm}
\newcommand{\labwidth}{0.092\linewidth}
\hspace{0.07\linewidth}
Task\hspace{\labwidth}
t = 1\hspace{\labwidth}
t = 3\hspace{\labwidth}
t = 5\hspace{\labwidth}
t = 7\hspace{\labwidth}
t = 9\hspace{\labwidth}
t = 11
   \caption{\small Examples of a robot using our model to successfully complete tool use tasks.  The robot must move the objects specified by the red symbols to the locations of the corresponding green symbols.  The robot uses a tool to simultaneously move several objects to their goal locations.}
\label{fig:robotic_control_qual}
\end{figure*}

%% file: sections/conclusion.tex
\section{Conclusion}

We present a method for learning predictive models from both passive observation and active interactions. Active interactions are usually more expensive and less readily-available than passive observation: for example, consider the amount of observational data of human activities on the internet. Active interaction, on the other hand, is especially difficult when the agent is trying to collect information about regions of the state-space which are difficult to reach. Without an existing policy that can guide the agent to those regions, time consuming on-policy exploration, expert teleoperated or kinesthetic demonstrations are often required, bringing additional costs. 

By learning a latent variable over the semi-observed actions, our approach is able to leverage passive observational data to improve action-conditioned predictive models, even in the presence of domain shift between observation and interaction data. Our experiments illustrate these benefits in two problem settings: driving and object manipulation, and find improvements both in prediction quality and in control performance when using these models for planning.

Overall, we hope that this work represents a first step towards enabling the use of broad, large-scale observational data when learning about the world.
However, limitations and open questions remain.
Our experiments studied a limited aspect of this broader problem where the observational data was either a different embodiment in the same environment (i.e. humans manipulating objects) or a different environment within the same underlying dataset (i.e. driving in Boston and Singapore). In practice, many source of passive observations will exhibit more substantial domain shift than those considered in this work. Hence, an important consideration for future work is to increase robustness to domain shift to realize greater benefits from using more large and diverse observational datasets. Finally, we focused our study on learning predictive models; an exciting direction for future work is to study how to incorporate similar forms of observational data in representation learning and reinforcement learning.

%% file: sections/acknowledgements.tex
\section*{Acknowledgements}
We would like to thank Kenneth Chaney for technical support.  We also like to thank Karl Pertsch and Drew Jaegle for insightful discussions.

This work was supported by the NSF GRFP,  ARL RCTA W911NF-10-2-0016, ARL DCIST CRA W911NF-17-2-0181, and by Honda Research Institute.

%% file: sections/appendix.tex
\clearpage
\begin{appendices}
\section{Full Architecture}
\begin{figure*}
    \centering
    \includegraphics[width=0.85\linewidth]{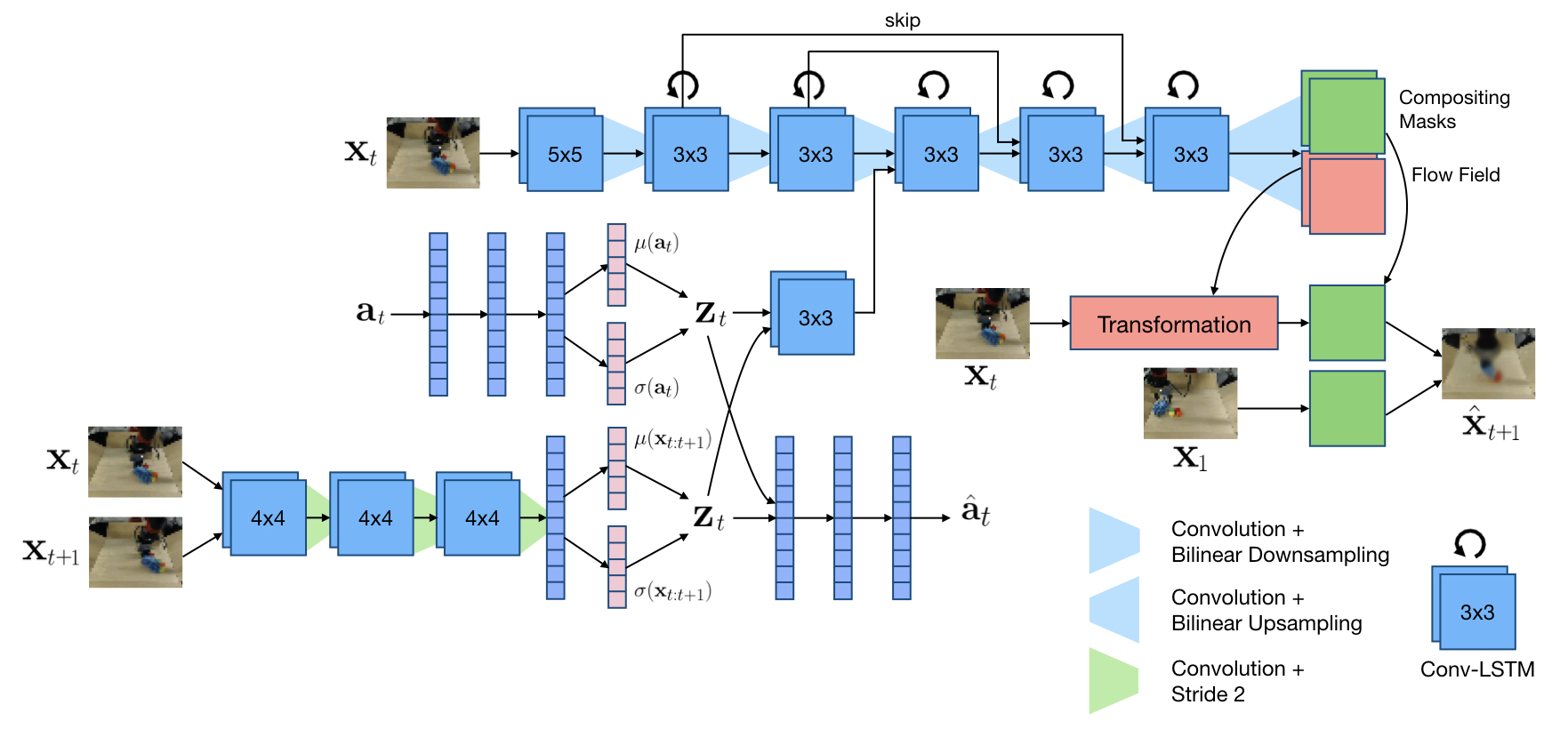}
    \caption{Our full architecture for learning from observation and interaction data. Our model is composed of the action encoder, inverse model, action decoder, and transition model. The action encoder and inverse model output distributions over $\textbf{z}_t$ conditioned on the action $\textbf{a}_t$ and image pair $\textbf{x}_t$}
    \label{fig:full_arch}
\end{figure*}

The full architecture of our model is presented in Figure~\ref{fig:full_arch}.
\label{appendix:full_architecture}

\section{Model Hyperparameters}
We selected our hyperparameters through cross-validation.  The hyperparameters that are shared between the domains are described in Table~\ref{tab:hyperparameters_all}.  The hyperparameters that are specific to the robotic manipulation domain are described in Table~\ref{tab:hyperparameters_robot}. The hyperparameters that are specific to the driving domain are described in Table~\ref{tab:hyperparameters_driving}.
\begin{table}[b]
\centering
\begin{tabular}{|l|c|}
\hline
    Hyperparameter & Value \\
\hline\hline
    Action decoder MSE weight & 0.0001 \\
    Action encoder KL weight & $10^-6$ \\
    Jensen-Shannon Divergence weight & $10^-7$ \\
    TV weight & 0.001 \\
    Image L1 reconstruction weight & 1.0 \\
    Optimizer & Adam \cite{Kingma2015} \\
    Learning rate & 0.001 \\
    Beta1 & 0.9 \\
    Beta2 & 0.999 \\
    Schedule sampling k & 900 \\
    Action encoder channels & 64 \\
    Action encoder layers & 3 \\
    Inverse model channels & 64 \\
    Inverse model layers & 3 \\
    Generator channels & 32 \\
\hline
\end{tabular}
\caption{Hyperparameter values}
\label{tab:hyperparameters_all}
\end{table}

\begin{table}
\centering
\begin{tabular}{|l|c|}
\hline
    Hyperparameter & Value \\
\hline\hline
    Dimensionality of $\textbf{z}^{\text{domain}}$ & 2 \\
    Dimensionality of $\textbf{z}^{\text{shared}}$ & 3 \\
    Prediction horizon & 15 \\
\hline
\end{tabular}
\caption{Hyperparameter values specific to the robotic manipulation domain}
\label{tab:hyperparameters_robot}
\end{table}

\begin{table}
\centering
\begin{tabular}{|l|c|}
\hline
    Hyperparameter & Value \\
\hline\hline
    Dimensionality of $\textbf{z}^{\text{domain}}$ & 0 \\
    Dimensionality of $\textbf{z}^{\text{shared}}$ & 3 \\
    Prediction horizon & 5 \\
\hline
\end{tabular}
\caption{Hyperparameter values specific to the driving domain}
\label{tab:hyperparameters_driving}
\end{table}

\section{Robot Planning and Control Experiments}
For the control experiments, each task is set up by placing one potential tool into the scene, as well as 2-3 objects to relocate which are specified to the planner by selecting start and goal pixels. The scenes are set up so that because the robot needs to move multiple objects, it is most effective for it to use the tool during its execution.
We present the hyperparameters used for the planner in our robotic control experiments in Table~\ref{tab:hparams_planning}. 
\begin{table}
\centering
\begin{tabular}{|l|c|}
\hline
    Hyperparameter & Value \\
\hline\hline
    Robot actions per trajectory & 20 \\
    Unique robot actions per trajectory & 6 (each repeated x3)\\
    CEM iterations & 4\\
    CEM candidate actions per iteration & 1200 \\
    CEM selection fraction & 0.05 \\
    Prediction horizon & 18 \\
    Number of goal-designating pixels & 3 \\
\hline

\end{tabular}
\caption{Hyperparameter values specific to the robot control experiments}
\label{tab:hparams_planning}
\end{table}

\section{Additional Implementation Details}
Additional implementation details are presented in this section.

\subsection{Batch Construction}
We constructed our batches so that they were made up of a fixed number of examples from each dataset.
In all of our experiments, we used a batch size of 12, made up of 9 samples from the interaction data and 3 samples from the observation data.

\subsection{Schedule Sampling}
In order to improve training, our system initially predicts images from the ground truth previous image.  As training continues, the system gradually shifts to using the predicted version of the previous frame.

The probability of sampling an image from the ground truth sequence is given by Equation \ref{eqn:schedule_sampling}.

\begin{equation}
    p = \min \left( \frac{k}{\exp{(i/k)}}, 1 \right)
    \label{eqn:schedule_sampling}
\end{equation}
The iteration number is i, while k is a hyperparameter that controls how many iterations it takes for the system to go from always using the ground truth images to always using the predicted images.  This sampling strategy was taken from \cite{lee2018savp}.

\section{Domain Shift}
\begin{figure}
    \centering
    \includegraphics[width=1.0\linewidth]{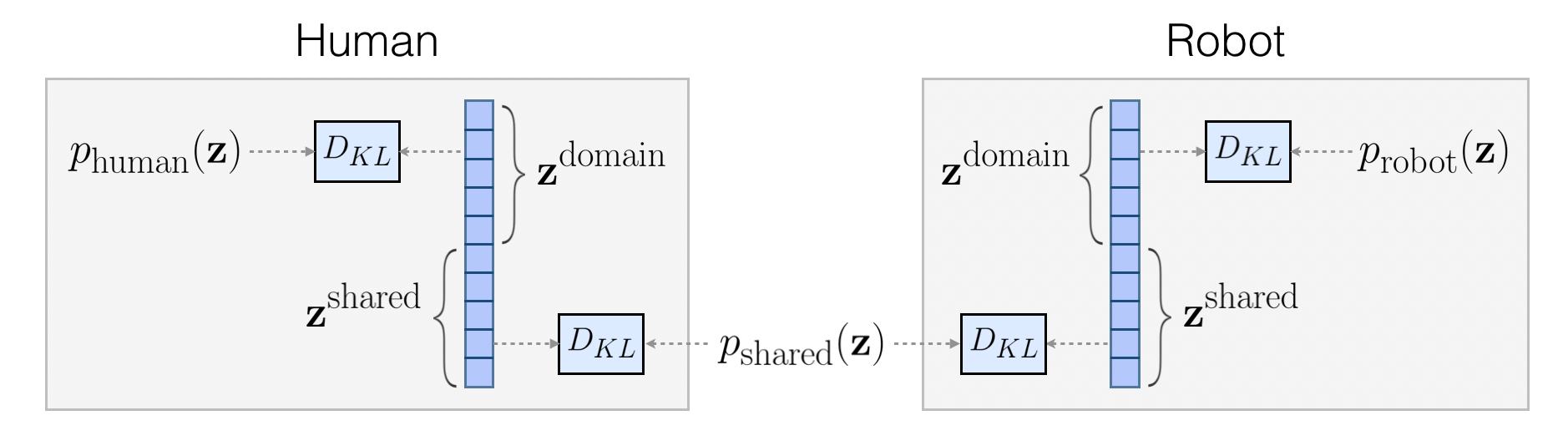}
    \caption{The partitioned latent space.  We partition our latent space $\z$ into two components, $\z^{\text{shared}}$, which captures the parts of the latent action that are shared between domains, and $\z^{\text{domain}}$, which captures the unique parts of the latent action.  We enforce this separation by learning the same prior for $\z^{\text{shared}}$ in all domains and a different prior for $\z^{\text{domain}}$ in each domain.}
    \label{fig:domain_shift}
\end{figure}

Our method of handling domain shift between datasets, described in Section~\ref{subsection:domain_shift}, is shown in Figure~\ref{fig:domain_shift}.

\section{Action Visualization}
We visualize the histogram of the robot actions in Figure \ref{fig:action_hists_2d}.  
All near-zero actions were removed from the histogram of the expert robot data to remove the long periods of time where the robot is stationary in that dataset.

\section{Additional Qualitative Results}
Additional qualitative results are presented in this section.
\subsection{Video Prediction in the Driving Domain}
A version of Figure~\ref{fig:nuscenes_predictions} with more images is shown in Figure~\ref{fig:nuscenes_predictions_full}.
We also present the sequence that is best for the baseline in Figure~\ref{fig:nuscenes_predictions_baseline_better}.
We present the sequence that has the median difference between methods in Figure~\ref{fig:nuscenes_predictions_middle}.

\input{sections/diagrams/nuscenes_predictions_appendix.tex}

\subsection{Video Prediction in the Robotic Manipulation Domain}
A version of Figure~\ref{fig:robotic_predictions} with more images is shown in Figure~\ref{fig:robotic_predictions_full}.
We also present the sequence that is best for the baseline in Figure~\ref{fig:robotic_predictions_baseline_better}.
We present the sequence that has the median difference between methods in Figure~\ref{fig:robotic_predictions_middle}.

\input{sections/diagrams/robot_predictions_appendix.tex}

\end{appendices}

%% file: sections/diagrams/nuscenes_predictions_appendix.tex
\begin{figure*}
\begin{center}

\rotatebox{90}{\scriptsize{Ground Truth}}
\includegraphics[width=0.13\linewidth]{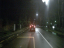}
\includegraphics[width=0.13\linewidth]{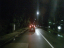}
\includegraphics[width=0.13\linewidth]{figures/nuscenes/traj_084/context/context_image_84_0_2.png}
\includegraphics[width=0.13\linewidth]{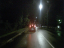}
\includegraphics[width=0.13\linewidth]{figures/nuscenes/traj_084/context/context_image_84_0_4.png}
\includegraphics[width=0.13\linewidth]{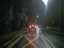}
\includegraphics[width=0.13\linewidth]{figures/nuscenes/traj_084/context/context_image_84_0_6.png}
\\
\rotatebox{90}{\scriptsize{Baseline}}
\includegraphics[width=0.13\linewidth]{figures/nuscenes/traj_084/context/context_image_84_0_0.png}
\includegraphics[width=0.13\linewidth]{figures/nuscenes/traj_084/context/context_image_84_0_1.png}
\includegraphics[width=0.13\linewidth]{figures/nuscenes/traj_084/baseline/gen_image_00084_00_00.png}
\includegraphics[width=0.13\linewidth]{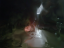}
\includegraphics[width=0.13\linewidth]{figures/nuscenes/traj_084/baseline/gen_image_00084_00_02.png}
\includegraphics[width=0.13\linewidth]{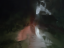}
\includegraphics[width=0.13\linewidth]{figures/nuscenes/traj_084/baseline/gen_image_00084_00_04.png}
\\
\rotatebox{90}{\scriptsize{POI (ours)}}
\includegraphics[width=0.13\linewidth]{figures/nuscenes/traj_084/context/context_image_84_0_0.png}
\includegraphics[width=0.13\linewidth]{figures/nuscenes/traj_084/context/context_image_84_0_1.png}
\includegraphics[width=0.13\linewidth]{figures/nuscenes/traj_084/ours/gen_image_00084_00_00.png}
\includegraphics[width=0.13\linewidth]{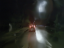}
\includegraphics[width=0.13\linewidth]{figures/nuscenes/traj_084/ours/gen_image_00084_00_02.png}
\includegraphics[width=0.13\linewidth]{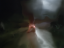}
\includegraphics[width=0.13\linewidth]{figures/nuscenes/traj_084/ours/gen_image_00084_00_04.png}

\end{center}
\vspace{-0.4cm}

\newcommand{\labwidth}{0.092\linewidth}
\hspace{0.07\linewidth}
t = -2\hspace{\labwidth}
t = -1\hspace{\labwidth}
t = 0\hspace{\labwidth}
t = 1\hspace{\labwidth}
t = 2\hspace{\labwidth}
t = 3\hspace{\labwidth}
t = 4

   \caption{\small Example predictions on the Singapore portion of the Nuscenes dataset. This sequence was selected for large MSE difference between the models.  We compare our model to the baseline of the SAVP model trained on the Boston data with actions. Our model is able to maintain the shape of the car in front.}
\label{fig:nuscenes_predictions_full}
\end{figure*}

\begin{figure*}
\begin{center}

\rotatebox{90}{\scriptsize{Ground Truth}}
\includegraphics[width=0.13\linewidth]{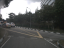}
\includegraphics[width=0.13\linewidth]{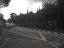}
\includegraphics[width=0.13\linewidth]{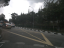}
\includegraphics[width=0.13\linewidth]{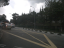}
\includegraphics[width=0.13\linewidth]{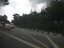}
\includegraphics[width=0.13\linewidth]{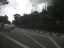}
\includegraphics[width=0.13\linewidth]{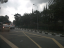}
\\
\rotatebox{90}{\scriptsize{Baseline}}
\includegraphics[width=0.13\linewidth]{figures/nuscenes/traj_052/context/context_image_52_0_0.png}
\includegraphics[width=0.13\linewidth]{figures/nuscenes/traj_052/context/context_image_52_0_1.png}
\includegraphics[width=0.13\linewidth]{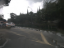}
\includegraphics[width=0.13\linewidth]{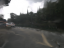}
\includegraphics[width=0.13\linewidth]{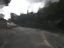}
\includegraphics[width=0.13\linewidth]{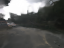}
\includegraphics[width=0.13\linewidth]{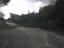}
\\
\rotatebox{90}{\scriptsize{POI (ours)}}
\includegraphics[width=0.13\linewidth]{figures/nuscenes/traj_052/context/context_image_52_0_0.png}
\includegraphics[width=0.13\linewidth]{figures/nuscenes/traj_052/context/context_image_52_0_1.png}
\includegraphics[width=0.13\linewidth]{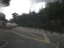}
\includegraphics[width=0.13\linewidth]{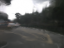}
\includegraphics[width=0.13\linewidth]{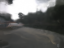}
\includegraphics[width=0.13\linewidth]{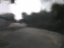}
\includegraphics[width=0.13\linewidth]{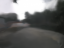}

\end{center}
\vspace{-0.4cm}

\newcommand{\labwidth}{0.092\linewidth}
\hspace{0.07\linewidth}
t = -2\hspace{\labwidth}
t = -1\hspace{\labwidth}
t = 0\hspace{\labwidth}
t = 1\hspace{\labwidth}
t = 2\hspace{\labwidth}
t = 3\hspace{\labwidth}
t = 4

   \caption{\small Example predictions on the Singapore portion of the Nuscenes dataset. This sequence was selected because the baseline had the largest improvement in MSE relative to our model.  We compare our model to the baseline of the SAVP model trained on the Boston data with actions.  Even in the worse case, our model performs comparably to the baseline model.}
\label{fig:nuscenes_predictions_baseline_better}
\end{figure*}

\begin{figure*}
\begin{center}

\rotatebox{90}{\scriptsize{Ground Truth}}
\includegraphics[width=0.13\linewidth]{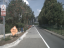}
\includegraphics[width=0.13\linewidth]{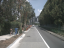}
\includegraphics[width=0.13\linewidth]{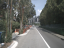}
\includegraphics[width=0.13\linewidth]{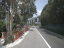}
\includegraphics[width=0.13\linewidth]{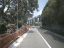}
\includegraphics[width=0.13\linewidth]{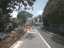}
\includegraphics[width=0.13\linewidth]{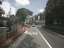}
\\
\rotatebox{90}{\scriptsize{Baseline}}
\includegraphics[width=0.13\linewidth]{figures/nuscenes/traj_038/context/context_image_38_0_0.png}
\includegraphics[width=0.13\linewidth]{figures/nuscenes/traj_038/context/context_image_38_0_1.png}
\includegraphics[width=0.13\linewidth]{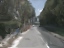}
\includegraphics[width=0.13\linewidth]{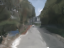}
\includegraphics[width=0.13\linewidth]{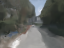}
\includegraphics[width=0.13\linewidth]{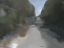}
\includegraphics[width=0.13\linewidth]{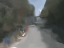}
\\
\rotatebox{90}{\scriptsize{POI (ours)}}
\includegraphics[width=0.13\linewidth]{figures/nuscenes/traj_038/context/context_image_38_0_0.png}
\includegraphics[width=0.13\linewidth]{figures/nuscenes/traj_038/context/context_image_38_0_1.png}
\includegraphics[width=0.13\linewidth]{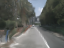}
\includegraphics[width=0.13\linewidth]{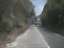}
\includegraphics[width=0.13\linewidth]{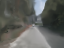}
\includegraphics[width=0.13\linewidth]{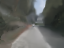}
\includegraphics[width=0.13\linewidth]{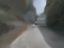}

\end{center}
\vspace{-0.4cm}

\newcommand{\labwidth}{0.092\linewidth}
\hspace{0.07\linewidth}
t = -2\hspace{\labwidth}
t = -1\hspace{\labwidth}
t = 0\hspace{\labwidth}
t = 1\hspace{\labwidth}
t = 2\hspace{\labwidth}
t = 3\hspace{\labwidth}
t = 4

   \caption{\small Example predictions on the Singapore portion of the Nuscenes dataset. This sequence was selected by ordering all of the sequences in the training set by the difference in MSE between the baseline and our model and selecting the middle sequence.  We compare our model to the baseline of the SAVP model trained on the Boston data with actions.  Even in the worse case, our model performs comparably to the baseline model.}
\label{fig:nuscenes_predictions_middle}
\end{figure*}

%% file: sections/diagrams/robot_predictions_appendix.tex
\begin{figure*}
\newcommand{\predwidth}{0.13\linewidth}
\begin{center}
\rotatebox{90}{\scriptsize{Ground Truth}}
\includegraphics[width=\predwidth]{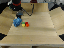}
\includegraphics[width=\predwidth]{figures/traj_00028_context/context_image_56_0_2.png}
\includegraphics[width=\predwidth]{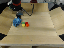}
\includegraphics[width=\predwidth]{figures/traj_00028_context/context_image_56_0_4.png}
\includegraphics[width=\predwidth]{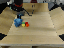}
\includegraphics[width=\predwidth]{figures/traj_00028_context/context_image_56_0_6.png}
\includegraphics[width=\predwidth]{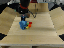} 
\\
\rotatebox{90}{\scriptsize{Baseline}}
\includegraphics[width=\predwidth]{figures/traj_00028_context/context_image_56_0_1.png}
\includegraphics[width=\predwidth]{figures/traj_00028_baseline/gen_image_00028_00_00.png}
\includegraphics[width=\predwidth]{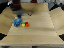}
\includegraphics[width=\predwidth]{figures/traj_00028_baseline/gen_image_00028_00_02.png}
\includegraphics[width=\predwidth]{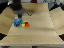}
\includegraphics[width=\predwidth]{figures/traj_00028_baseline/gen_image_00028_00_04.png}
\includegraphics[width=\predwidth]{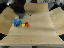}
\\
\rotatebox{90}{\scriptsize{POI (ours)}}
\includegraphics[width=\predwidth]{figures/traj_00028_context/context_image_56_0_1.png}
\includegraphics[width=\predwidth]{figures/traj_00028_ours/gen_image_00028_00_00.png}
\includegraphics[width=\predwidth]{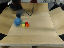}
\includegraphics[width=\predwidth]{figures/traj_00028_ours/gen_image_00028_00_02.png}
\includegraphics[width=\predwidth]{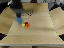}
\includegraphics[width=\predwidth]{figures/traj_00028_ours/gen_image_00028_00_04.png}
\includegraphics[width=\predwidth]{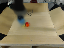}
\\
\newcommand{\labwidth}{0.092\linewidth}
\hspace{0.07\linewidth}
t = -1\hspace{\labwidth}
t = 0\hspace{\labwidth}
t = 1\hspace{\labwidth}
t = 2\hspace{\labwidth}
t = 3\hspace{\labwidth}
t = 4\hspace{\labwidth}
t = 5
\vspace{0.4cm}
\\
\rotatebox{90}{\scriptsize{Ground Truth}}
\includegraphics[width=\predwidth]{figures/traj_00028_context/context_image_56_0_8.png}
\includegraphics[width=\predwidth]{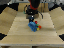}
\includegraphics[width=\predwidth]{figures/traj_00028_context/context_image_56_0_10.png}
\includegraphics[width=\predwidth]{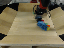}
\includegraphics[width=\predwidth]{figures/traj_00028_context/context_image_56_0_12.png}
\includegraphics[width=\predwidth]{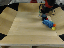}
\includegraphics[width=\predwidth]{figures/traj_00028_context/context_image_56_0_14.png}
\\
\rotatebox{90}{\scriptsize{Baseline}}
\includegraphics[width=\predwidth]{figures/traj_00028_baseline/gen_image_00028_00_06.png}
\includegraphics[width=\predwidth]{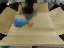}
\includegraphics[width=\predwidth]{figures/traj_00028_baseline/gen_image_00028_00_08.png}
\includegraphics[width=\predwidth]{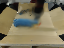}
\includegraphics[width=\predwidth]{figures/traj_00028_baseline/gen_image_00028_00_10.png}
\includegraphics[width=\predwidth]{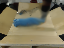}
\includegraphics[width=\predwidth]{figures/traj_00028_baseline/gen_image_00028_00_12.png}
\\
\rotatebox{90}{\scriptsize{POI (ours)}}
\includegraphics[width=\predwidth]{figures/traj_00028_ours/gen_image_00028_00_06.png}
\includegraphics[width=\predwidth]{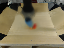}
\includegraphics[width=\predwidth]{figures/traj_00028_ours/gen_image_00028_00_08.png}
\includegraphics[width=\predwidth]{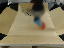}
\includegraphics[width=\predwidth]{figures/traj_00028_ours/gen_image_00028_00_10.png}
\includegraphics[width=\predwidth]{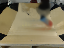}
\includegraphics[width=\predwidth]{figures/traj_00028_ours/gen_image_00028_00_12.png}
\\

\end{center}
\vspace{-0.4cm}
\newcommand{\labwidth}{0.092\linewidth}
\hspace{0.07\linewidth}
t = 6\hspace{\labwidth}
t = 7\hspace{\labwidth}
t = 8\hspace{\labwidth}
t = 9\hspace{\labwidth}
t = 10\hspace{\labwidth}
t = 11\hspace{\labwidth}
t = 12

   \caption{\small Example predictions on the robotic dataset. The first image is the context image.  We compare our model to the baseline of the SAVP model trained with random robot data. This sequence was selected to maximize the MSE difference between the models. Our model more accurately predicts both the tool and the object it pushes.}
\label{fig:robotic_predictions_full}
\end{figure*}

%%%%%%%%%%%%%%%%%%%%%%%%%%%%%%%%%%%%%%%%%%%%%%%%%%%%%%%%%%
\begin{figure*}
\newcommand{\predwidth}{0.13\linewidth}
\begin{center}
\rotatebox{90}{\scriptsize{Ground Truth}}
\includegraphics[width=\predwidth]{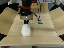}
\includegraphics[width=\predwidth]{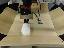}
\includegraphics[width=\predwidth]{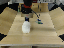}
\includegraphics[width=\predwidth]{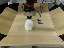}
\includegraphics[width=\predwidth]{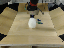}
\includegraphics[width=\predwidth]{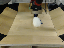}
\includegraphics[width=\predwidth]{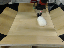} 
\\
\rotatebox{90}{\scriptsize{Baseline}}
\includegraphics[width=\predwidth]{figures/robot_traj_00109/context/context_image_109_0_1.png}
\includegraphics[width=\predwidth]{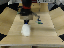}
\includegraphics[width=\predwidth]{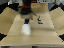}
\includegraphics[width=\predwidth]{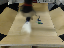}
\includegraphics[width=\predwidth]{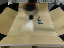}
\includegraphics[width=\predwidth]{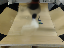}
\includegraphics[width=\predwidth]{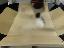}
\\
\rotatebox{90}{\scriptsize{POI (ours)}}
\includegraphics[width=\predwidth]{figures/robot_traj_00109/context/context_image_109_0_1.png}
\includegraphics[width=\predwidth]{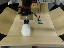}
\includegraphics[width=\predwidth]{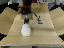}
\includegraphics[width=\predwidth]{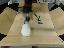}
\includegraphics[width=\predwidth]{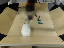}
\includegraphics[width=\predwidth]{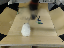}
\includegraphics[width=\predwidth]{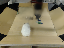}
\\
\newcommand{\labwidth}{0.092\linewidth}
\hspace{0.07\linewidth}
t = -1\hspace{\labwidth}
t = 0\hspace{\labwidth}
t = 1\hspace{\labwidth}
t = 2\hspace{\labwidth}
t = 3\hspace{\labwidth}
t = 4\hspace{\labwidth}
t = 5
\vspace{0.4cm}
\\
\rotatebox{90}{\scriptsize{Ground Truth}}
\includegraphics[width=\predwidth]{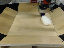}
\includegraphics[width=\predwidth]{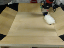}
\includegraphics[width=\predwidth]{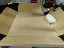}
\includegraphics[width=\predwidth]{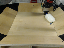}
\includegraphics[width=\predwidth]{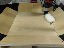}
\includegraphics[width=\predwidth]{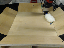}
\includegraphics[width=\predwidth]{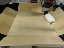}
\\
\rotatebox{90}{\scriptsize{Baseline}}
\includegraphics[width=\predwidth]{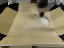}
\includegraphics[width=\predwidth]{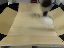}
\includegraphics[width=\predwidth]{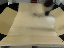}
\includegraphics[width=\predwidth]{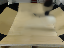}
\includegraphics[width=\predwidth]{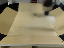}
\includegraphics[width=\predwidth]{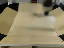}
\includegraphics[width=\predwidth]{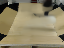}
\\
\rotatebox{90}{\scriptsize{POI (ours)}}
\includegraphics[width=\predwidth]{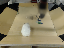}
\includegraphics[width=\predwidth]{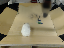}
\includegraphics[width=\predwidth]{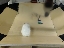}
\includegraphics[width=\predwidth]{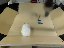}
\includegraphics[width=\predwidth]{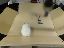}
\includegraphics[width=\predwidth]{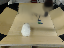}
\includegraphics[width=\predwidth]{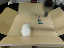}
\\

\end{center}
\vspace{-0.4cm}
\newcommand{\labwidth}{0.092\linewidth}
\hspace{0.07\linewidth}
t = 6\hspace{\labwidth}
t = 7\hspace{\labwidth}
t = 8\hspace{\labwidth}
t = 9\hspace{\labwidth}
t = 10\hspace{\labwidth}
t = 11\hspace{\labwidth}
t = 12

   \caption{\small Example predictions on the robotic dataset. The first image is the context image.  We compare our model to the baseline of the SAVP model trained with random robot data. This sequence was selected to maximize so that the baseline had the largest improvement in MSE relative to our model. Our model fails because it was too pessimistic about grasping the narrow handle of the brush.}
\label{fig:robotic_predictions_baseline_better}
\end{figure*}

%%%%%%%%%%%%%%%%%%%%%%%%%%%%%%%%%%%%%%%%%%%%%%%%%%%%%%

\begin{figure*}
\newcommand{\predwidth}{0.13\linewidth}
\begin{center}
\rotatebox{90}{\scriptsize{Ground Truth}}
\includegraphics[width=\predwidth]{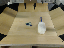}
\includegraphics[width=\predwidth]{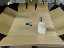}
\includegraphics[width=\predwidth]{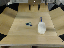}
\includegraphics[width=\predwidth]{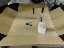}
\includegraphics[width=\predwidth]{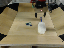}
\includegraphics[width=\predwidth]{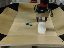}
\includegraphics[width=\predwidth]{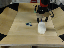} 
\\
\rotatebox{90}{\scriptsize{Baseline}}
\includegraphics[width=\predwidth]{figures/robot_traj_00080/context/context_image_80_0_1.png}
\includegraphics[width=\predwidth]{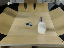}
\includegraphics[width=\predwidth]{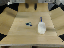}
\includegraphics[width=\predwidth]{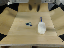}
\includegraphics[width=\predwidth]{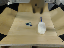}
\includegraphics[width=\predwidth]{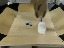}
\includegraphics[width=\predwidth]{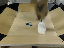}
\\
\rotatebox{90}{\scriptsize{POI (ours)}}
\includegraphics[width=\predwidth]{figures/robot_traj_00080/context/context_image_80_0_1.png}
\includegraphics[width=\predwidth]{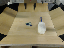}
\includegraphics[width=\predwidth]{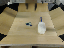}
\includegraphics[width=\predwidth]{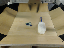}
\includegraphics[width=\predwidth]{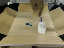}
\includegraphics[width=\predwidth]{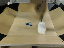}
\includegraphics[width=\predwidth]{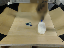}
\\
\newcommand{\labwidth}{0.092\linewidth}
\hspace{0.07\linewidth}
t = -1\hspace{\labwidth}
t = 0\hspace{\labwidth}
t = 1\hspace{\labwidth}
t = 2\hspace{\labwidth}
t = 3\hspace{\labwidth}
t = 4\hspace{\labwidth}
t = 5
\vspace{0.4cm}
\\
\rotatebox{90}{\scriptsize{Ground Truth}}
\includegraphics[width=\predwidth]{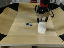}
\includegraphics[width=\predwidth]{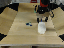}
\includegraphics[width=\predwidth]{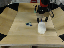}
\includegraphics[width=\predwidth]{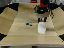}
\includegraphics[width=\predwidth]{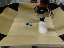}
\includegraphics[width=\predwidth]{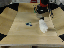}
\includegraphics[width=\predwidth]{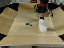}
\\
\rotatebox{90}{\scriptsize{Baseline}}
\includegraphics[width=\predwidth]{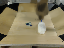}
\includegraphics[width=\predwidth]{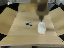}
\includegraphics[width=\predwidth]{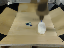}
\includegraphics[width=\predwidth]{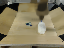}
\includegraphics[width=\predwidth]{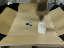}
\includegraphics[width=\predwidth]{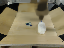}
\includegraphics[width=\predwidth]{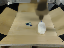}
\\
\rotatebox{90}{\scriptsize{POI (ours)}}
\includegraphics[width=\predwidth]{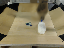}
\includegraphics[width=\predwidth]{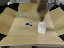}
\includegraphics[width=\predwidth]{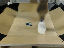}
\includegraphics[width=\predwidth]{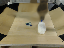}
\includegraphics[width=\predwidth]{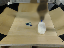}
\includegraphics[width=\predwidth]{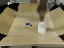}
\includegraphics[width=\predwidth]{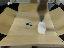}
\\

\end{center}
\vspace{-0.4cm}
\newcommand{\labwidth}{0.092\linewidth}
\hspace{0.07\linewidth}
t = 6\hspace{\labwidth}
t = 7\hspace{\labwidth}
t = 8\hspace{\labwidth}
t = 9\hspace{\labwidth}
t = 10\hspace{\labwidth}
t = 11\hspace{\labwidth}
t = 12

   \caption{\small Example predictions on the robotic dataset. The first image is the context image.  We compare our model to the baseline of the SAVP model trained with random robot data. This sequence had the median difference in MSE between our model and the baseline.}
\label{fig:robotic_predictions_middle}
\end{figure*}